\documentclass{article}
\usepackage[preprint]{neurips_2022}
\usepackage[utf8]{inputenc}
\usepackage[T1]{fontenc}
\usepackage[english]{babel}
\usepackage[colorlinks]{hyperref}
\usepackage{url}
\usepackage{amsfonts}
\usepackage{nicefrac}
\usepackage{microtype}
\usepackage{amsmath}
\usepackage{amssymb}
\usepackage{mathtools}
\usepackage{subcaption}
\usepackage{ragged2e}
\usepackage{tikz}
\usepackage{stackengine}
\usepackage{etoolbox}
\usepackage{xspace}
\usepackage{xpatch}
\usepackage{cuted}
\usepackage{enumerate}
\usepackage{xstring}
\usepackage{natbib}
\usepackage{setspace}
\usepackage{makecell}
\usepackage{graphicx}
\usepackage{changepage}
\usepackage{enumitem}
\usepackage{cuted}
\usepackage{titlesec}
\usepackage{cancel}
\usepackage{eqparbox}
\usepackage{wrapfig}
\usepackage[hang,flushmargin,bottom]{footmisc}
\usepackage[capitalise,noabbrev,nameinlink]{cleveref}
\usepackage[useregional=numeric]{datetime2}
\usepackage[linesnumbered,ruled,noend]{algorithm2e}
\usepackage{pifont}
\usepackage{listings}
\usepackage{titlecaps}
\usepackage{environ}
\usepackage{tabularray}
\usepackage{titletoc}
\usepackage{tocbasic}

\definecolor{darkgreen}{RGB}{120, 166, 90}
\definecolor{darkblue}{RGB}{80, 131, 193}
\definecolor{darkred}{RGB}{182, 42, 21}
\definecolor{darkorange}{RGB}{218, 149, 75}

\usetikzlibrary{positioning,arrows.meta,fit,calc,backgrounds}
\tikzset{%
node distance=2em, auto,
every node/.style={line width=0.7pt},
det/.style={draw=black, rectangle, minimum size=2.5em, inner sep=0.1ex},
lat/.style={draw=black, circle, minimum size=2.5em, inner sep=0.1ex},
obs/.style={draw=black, circle, fill=black!15, minimum size=2.5em, inner sep=0.1ex},
fac/.style={draw=black, rectangle, fill=black, minimum size=.6em, inner sep=0em},
dummy/.style={draw=none, circle, minimum size=2.5em},
plate/.style={draw=black, rounded corners, inner sep=.8em, yshift=-.7em, align=right},
box/.style={draw=black, rounded corners, inner sep=.4em, align=center},
generates/.style={->, -{Stealth[length=.6em, inset=0pt]}, line width=0.7pt},
undirected/.style={line width=0.7pt},
}

\setlist[itemize]{leftmargin=1.5em,itemsep=.1em,topsep=.1em}
\titlespacing*{\paragraph}{0pt}{0ex plus .1ex}{1em}
\DeclareTOCStyleEntry[beforeskip=.2em plus 1pt]{tocline}{section}
\xapptocmd\normalsize{%
\abovedisplayskip=.8em plus .2em minus .2em
\belowdisplayskip=.6em plus .1em minus .1em
\abovedisplayshortskip=.8em plus .2em minus .2em
\belowdisplayshortskip=.6em plus .1em minus .1em
}{}{}
\newcommand{\itempar}[1]{\item\textbf{#1}\quad}

\newcommand\blfootnote[1]{%
\begingroup
\renewcommand\thefootnote{}\footnote{#1}%
\addtocounter{footnote}{-1}%
\endgroup}

\setcitestyle{authoryear,round,citesep={;},aysep={,},yysep={;}}
\renewcommand{\cite}[1]{\citep{#1}}

\creflabelformat{equation}{#2\textup{#1}#3}
\crefname{algocf}{Algorithm}{Algorithms}
\Crefname{algocf}{Algorithm}{Algorithms}

\definecolor{mydarkblue}{rgb}{0,0.08,0.45}
\hypersetup{%
colorlinks=true,
linkcolor=mydarkblue,
citecolor=mydarkblue,
filecolor=mydarkblue,
urlcolor=mydarkblue}

\graphicspath{{figures/}}

\makeatletter
\def\hlinewd#1{%
\noalign{\ifnum0=`}\fi\hrule \@height #1 \futurelet
\reserved@a\@xhline}
\makeatother
\UseTblrLibrary{booktabs}
\NewColumnType{L}[1]{Q[l,#1]}
\NewColumnType{C}[1]{Q[c,#1]}
\NewColumnType{R}[1]{Q[r,#1]}

\NewEnviron{mytabular}[1]{%
  \setlength\heavyrulewidth{1.2pt}
  \setlength\lightrulewidth{.5pt}
  \setlength\aboverulesep{.4ex}%
  \setlength\belowrulesep{.8ex}%
  \begin{booktabs}[expand=\BODY]{#1}
    \BODY
  \end{booktabs}
}

\makeatletter
\DeclareDocumentCommand\todo{g}{%
\def\@message{\IfNoValueTF{#1}{TODO}{TODO: #1}}
\textbf{\textcolor[HTML]{FF8811}{\@message}}
\@latex@warning{\@message}{}{}}
\makeatother

\makeatletter
\newcommand{\ProcessDigit}[1]{%
  \ifnum\lst@mode=\lst@Pmode\relax{\color[HTML]{005cc5} #1}\else #1 \fi}
\makeatother
\makeatletter
\newcommand{\replunderscores}[1]{\expandafter\@repl@underscores#1_\relax}
\def\@repl@underscores#1_#2\relax{%
  \ifx \relax #2\relax #1%
  \else #1\textunderscore\@repl@underscores#2\relax \fi}
\makeatother

\renewcommand\d{\mathop{}\!\textnormal{\slshape d}}

\makeatletter
\newcommand{\removeParBefore}{\ifvmode\vspace*{-\baselineskip}\setlength{\parskip}{0ex}\fi}
\newcommand{\removeParAfter}{\@ifnextchar\par\@gobble\relax}
\newcommand{\eq}{\begingroup\removeParBefore\endlinechar=32 \eqinner}
\newcommand{\eqinner}[2][aligned]{\endlinechar=32%
\begin{gather}\begin{#1}#2\end{#1}\end{gather}\endgroup\removeParAfter}
\makeatother

\DeclareDocumentCommand{\p}{ D<>{p} D<>{} r() }{
\def\content{#3}\patchcmd{\content}{|}{\;#2\vert\;}{}{}
\ensuremath{#1 #2(\content #2)}}

\DeclareDocumentCommand{\P}{ D<>{P} D<>{\big} r() }{
\def\content{#3}\patchcmd{\content}{|}{\;#2\vert\;}{}{}
\ensuremath{\operatorname{#1}#2(\content #2)}}

\DeclareDocumentCommand{\E}{ D<>{E} E{_}{{}} D<>{\big} r[] }{
\def\content{#4}\patchcmd{\content}{|}{\;#3\vert\;}{}{}
\ensuremath{\operatorname{#1}_{#2}#3[\content #3]}}

\DeclareDocumentCommand{\D}{ D<>{D} D<>{\big} r[] }{
\def\content{#3}\patchcmd{\content}{||}{\;#2\|\;}{}{}
\ensuremath{\operatorname{#1}\!#2[\content #2]}}

\NewDocumentCommand{\Nor}{ r() }{\P<\mathrm{Normal}\!>(#1)}
\NewDocumentCommand{\Cat}{ r() }{\P<\mathrm{Cat}\!>(#1)}
\NewDocumentCommand{\Bin}{ r() }{\P<\mathrm{Bin}\!>(#1)}
\NewDocumentCommand{\Bet}{ r() }{\P<\mathrm{Beta}\!>(#1)}
\NewDocumentCommand{\Ber}{ r() }{\P<\mathrm{Bernoulli}\!>(#1)}
\NewDocumentCommand{\Dir}{ r() }{\P<\mathrm{Dir}\!>(#1)}
\NewDocumentCommand{\VecCat}{ r() }{\P<\mathrm{VecCat}\!>(#1)}

\DeclareDocumentCommand{\KL}{ D<>{\big} r[] }{\D<KL><#1>[#2]}
\DeclareDocumentCommand{\H}{ D<>{\big} r[] }{\E<H><#1>[#2]}
\DeclareDocumentCommand{\I}{ D<>{\big} r[] }{\E<I><#1>[#2]}

\DeclareDocumentCommand{\lnpp}{ D<>{} r() }{\ensuremath{\p<\ln p_\phi><#1>(#2)}}
\DeclareDocumentCommand{\pp}{ D<>{} r() }{\ensuremath{\p<p_\phi><#1>(#2)}}
\DeclareDocumentCommand{\qp}{ D<>{} r() }{\ensuremath{\p<q_\phi><#1>(#2)}}

\DeclareDocumentCommand{\SymLogNormal}{ D<>{} r() }{
\ensuremath{\p<\operatorname{SymLogNormal}><#1>(#2)}}

\DeclareDocumentCommand{\enc}{ D<>{} r() }{\ensuremath{\p<\mathrm{enc}><#1>(#2)}}
\DeclareDocumentCommand{\dec}{ D<>{} r() }{\ensuremath{\p<\mathrm{dec}><#1>(#2)}}
\DeclareDocumentCommand{\mgr}{ D<>{} r() }{\ensuremath{\p<\mathrm{mgr}><#1>(#2)}}
\DeclareDocumentCommand{\wkr}{ D<>{} r() }{\ensuremath{\p<\mathrm{wkr}><#1>(#2)}}

\DeclareDocumentCommand{\reprW}{ D<>{} r() }{\ensuremath{\p<\mathrm{repr}_\theta><#1>(#2)}}
\DeclareDocumentCommand{\dynW}{ D<>{} r() }{\ensuremath{\p<\mathrm{dyn}_\theta><#1>(#2)}}
\DeclareDocumentCommand{\recW}{ D<>{} r() }{\ensuremath{\p<\mathrm{rec}_\theta><#1>(#2)}}
\DeclareDocumentCommand{\rewW}{ D<>{} r() }{\ensuremath{\p<\mathrm{rew}_\theta><#1>(#2)}}

\DeclareDocumentCommand{\encW}{ D<>{} r() }{\ensuremath{\p<\mathrm{enc}_\phi><#1>(#2)}}
\DeclareDocumentCommand{\decW}{ D<>{} r() }{\ensuremath{\p<\mathrm{dec}_\phi><#1>(#2)}}
\DeclareDocumentCommand{\mgrW}{ D<>{} r() }{\ensuremath{\p<\mathrm{mgr}_\psi><#1>(#2)}}
\DeclareDocumentCommand{\wkrW}{ D<>{} r() }{\ensuremath{\p<\mathrm{wkr}_\xi><#1>(#2)}}

\title{Deep Hierarchical Planning from Pixels}

\author{
  Danijar Hafner\,\textsuperscript{\normalfont 1 2 3} \\ \And
  Kuang-Huei Lee\,\textsuperscript{\normalfont 2} \\ \And
  Ian Fischer\,\textsuperscript{\normalfont 2} \\ \And
  Pieter Abbeel\,\textsuperscript{\normalfont 1 4} \\
}

\begin{document}

\blfootnote{
\textsuperscript{1}UC Berkeley \enskip
\textsuperscript{2}Google Research \enskip
\textsuperscript{3}University of Toronto \enskip
\textsuperscript{4}Covariant \enskip \\
Correspondence to: Danijar Hafner <mail@danijar.com>.
Preprint. Under review. \\
Project website with videos and code: \url{https://danijar.com/director}
}

\vspace*{-2ex}
\maketitle

\begin{abstract}
\begin{hyphenrules}{nohyphenation}
Intelligent agents need to select long sequences of actions to solve complex tasks. While humans easily break down tasks into subgoals and reach them through millions of muscle commands, current artificial intelligence is limited to tasks with horizons of a few hundred decisions, despite large compute budgets. Research on hierarchical reinforcement learning aims to overcome this limitation but has proven to be challenging, current methods rely on manually specified goal spaces or subtasks, and no general solution exists. We introduce Director, a practical method for learning hierarchical behaviors directly from pixels by planning inside the latent space of a learned world model. The high-level policy maximizes task and exploration rewards by selecting latent goals and the low-level policy learns to achieve the goals. Despite operating in latent space, the decisions are interpretable because the world model can decode goals into images for visualization. Director outperforms exploration methods on tasks with sparse rewards, including 3D maze traversal with a quadruped robot from an egocentric camera and proprioception, without access to the global position or top-down view that was used by prior work. Director also learns successful behaviors across a wide range of environments, including visual control, Atari games, and DMLab levels.

\end{hyphenrules}
\end{abstract}

\begin{figure}[b!]
\centering
\includegraphics[width=.17\textwidth]{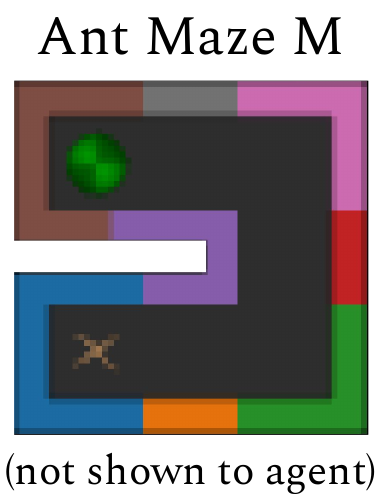}\hfill%
\includegraphics[width=.82\textwidth]{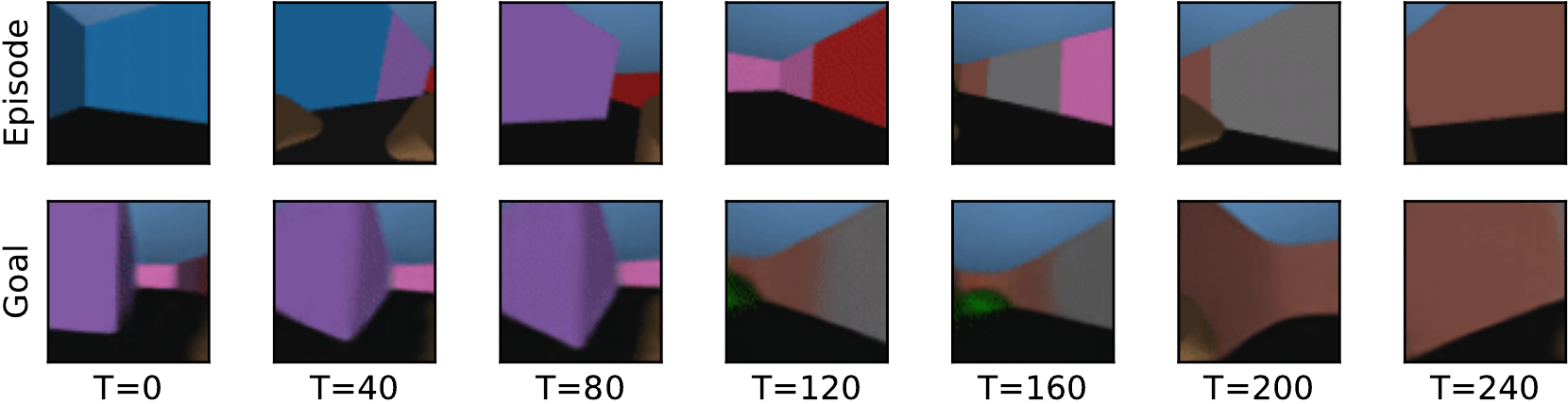}%
\caption{%
Director on Ant Maze M from egocentric camera inputs. 
The top row shows agent inputs. 
The bottom row shows the internal subgoals of the agent. 
The goals are latent vectors that Director's world model can decode into images for human inspection. 
Director solves this sparse reward task by breaking it down into internal goals. 
It first targets the purple wall in the middle of the maze. 
Once reached, it targets the reward object, and then the brown wall behind it to step onto the reward object.
}
\label{fig:first}
\end{figure}

\section{Introduction}
\label{sec:intro}

Artificial agents have achieved remarkable performance on reactive video games \citep{mnih2015dqn,badia2020agent57} or board games that last for a few hundred moves \citep{silver2017alphago}.
However, solving complex control problems can require millions of time steps. For example, consider a robot that needs to navigate along the sidewalk and cross streets to buy groceries and then return home and cook a meal with those groceries. 
Manually specifying subtasks or dense rewards for such complex tasks would not only be expensive but also prone to errors and require tremendous effort to capture special cases \citep{chen2021dvd,ahn2022saycan}.
Even training a robot to simply walk forward can require specifying ten different reward terms \citep{kumar2021rma}, making reward engineering a critical component of such systems. 
Humans naturally break long tasks into subgoals, each of which is easy to achieve. 
In contrast, most current reinforcement learning algorithms reason purely at the clock rate of their primitive actions. 
This poses a key bottleneck of current reinforcement learning methods that could be challenging to solve by simply increasing the computational budget.

Hierarchical reinforcement learning (HRL) \citep{dayan1992feudal,parr1997ham,sutton1999options} aims to automatically break long-horizon tasks into subgoals or commands that are easier to achieve, typically by learning high-level controllers that operate at more abstract time scales and provide commands to low-level controllers that select primitive actions. 
However, most HRL approaches require domain knowledge to break down tasks, either through manually specified subtasks \citep{tessler2017dsn} or semantic goal spaces such as global XY coordinates for navigation tasks \citep{andrychowicz2017her,nachum2018hiro} or robot poses \citep{gehring2021hsd3}. 
Attempts at learning hierarchies directly from sparse rewards have had limited success \citep{vezhnevets2017fun} and required providing task reward to the low-level controller, calling into question the benefit of their high-level controller.

In this paper, we present Director, a practical method for learning hierarchical behaviors directly from pixels by planning inside the latent space of a learned world model. 
We observe the effectiveness of Director on long-horizon tasks with very sparse rewards and demonstrate its generality by learning successfully in a wide range of domains. 
The key insights of Director are to leverage the representations of the world model, select goals in a compact discrete space to aid learning for the high-level policy, and to use a simple form of temporally-extended exploration in the high-level policy.

\paragraph{Contributions}

The key contributions of this paper are summarized as follows:

\begin{itemize}
\item We describe a practical, general, and interpretable algorithm for learning hierarchical behaviors within a world model trained from pixels, which we call Director (\Cref{sec:method}).
\item We introduce two sparse reward benchmarks that underscore the limitations of traditional flat RL approaches and find that Director solves these challenging tasks (\Cref{sec:sparse}).
\item We demonstrate that Director successfully learns in a wide range of traditional RL environments, including Atari, Control Suite, DMLab, and Crafter (\Cref{sec:standard}).
\item We visualize the latent goals that Director selects for breaking down various tasks, providing insights into its decision making (\Cref{sec:interp}).
\end{itemize}
\begin{figure}[t]
\vspace*{-5ex}
\centering
\includegraphics[width=\textwidth]{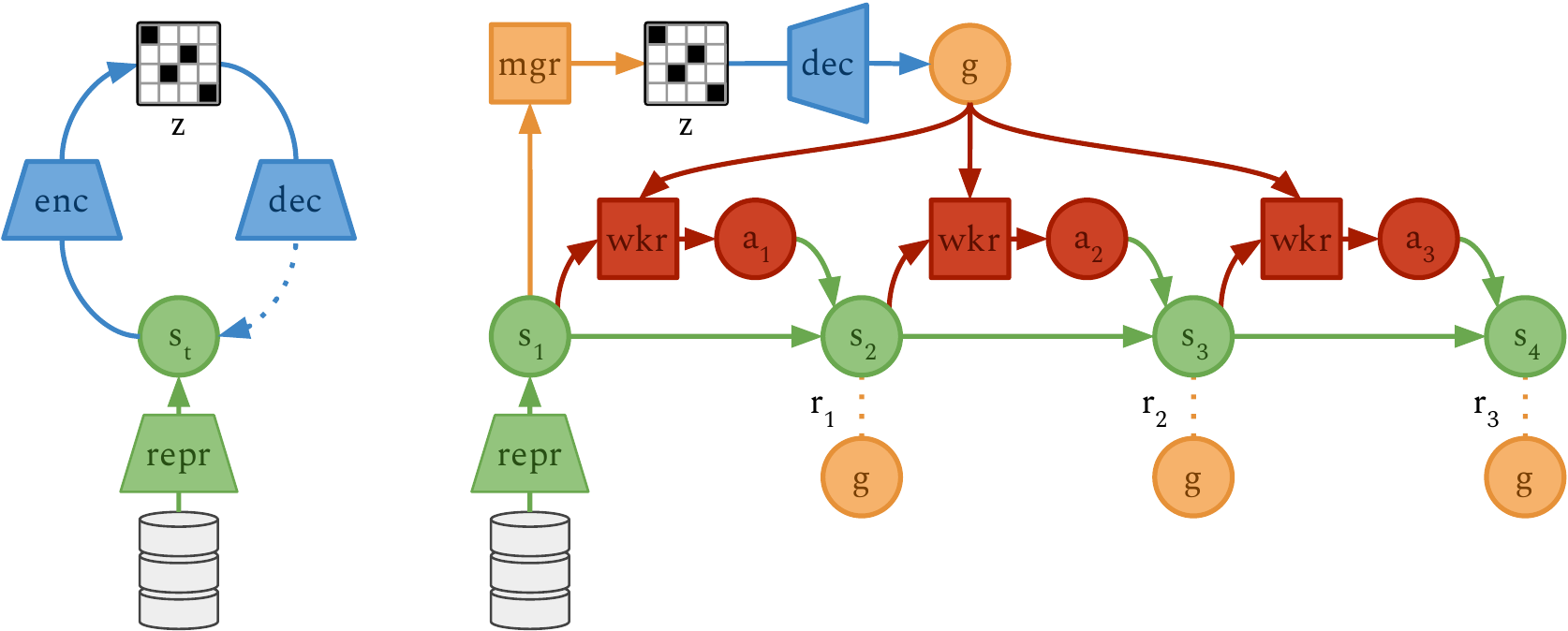}
\caption{%
Director is based on the \textcolor{darkgreen}{\textbf{world model}} of PlaNet \citep{hafner2018planet} that predicts ahead in a compact representation space. The world model is trained by reconstructing images using a neural network not shown in the figure. Director then learns three additional components, which are all optimized concurrently.
On the left, the \textcolor{darkblue}{\textbf{goal autoencoder}} compresses the feature vectors $s_t$ into vectors of discrete codes $z\sim\enc(z|s_t)$.
On the right, the \textcolor{darkorange}{\textbf{manager}} policy $\mgr(z|s_t)$ selects abstract actions in this discrete space every $K=8$ steps, which the goal decoder then turns into feature space goals $g=\dec(z)$. 
The \textcolor{darkred}{\textbf{worker}} policy $\wkr(a_t|s_t,g)$ receives the current feature vector and goal as input to decide primitive actions that maximize the similarity rewards $r_t$ to the goal. 
The manager maximizes the task reward and an exploration bonus based on the autoencoder reconstruction error, implementing temporally-extended exploration.
}
\label{fig:method}
\end{figure}

\section{Director}
\label{sec:method}

Director is a reinforcement learning algorithm that learns hierarchical behaviors directly from pixels. 
As shown in \cref{fig:method}, Director learns a world model for representation learning and planning, a goal autoencoder that discretizes the possible goals to make them easier for the manager to choose, a manager policy that selects goals every fixed number of steps to maximize task and exploration rewards, and a worker policy that learns to reach the goals through primitive actions.
All components are optimized throughout the course of learning by performing one gradient step every fixed number of environment steps. 
The world model is trained from a replay buffer whereas the goal autoencoder is trained on the world model representations and the policies are optimized from imagined rollouts.
For the pseudo code of Director, refer to \cref{sec:algo}.

\subsection{World Model}
\label{sec:wm}

Director learns a world model that compresses the history of observations into a compact feature space and enables planning in this space \citep{watter2015e2c,zhang2018solar}.
We use the Recurrent State Space Model (RSSM) model of PlaNet \citep{hafner2018planet}, which we briefly review here to introduce notation. 
The world model consists of four neural networks that are optimized jointly:

\eq{
&\text{Model representation:} &&\reprW(s_t|s_{t-1},a_{t-1},x_t) \quad
&&\text{Model decoder:} &&\recW(s_t) \approx x_t \\
&\text{Model dynamics:} &&\dynW(s_t|s_{t-1},a_{t-1}) \quad
&&\text{Reward predictor:} &&\rewW(s_{t+1}) \approx r_t \\
}

The representation model integrates actions $a_t$ and observations $x_t$ into the latent states $s_t$.
The dynamics model predicts future states without the corresponding observations.
The decoder reconstructs observations to provide a rich learning signal.
The reward predictor later allows learning policies by planning in the compact latent space, without decoding images. 
The world model is optimized end-to-end on subsequences from the replay buffer by stochastic gradient descent on the variational objective \citep{hinton1993vi,kingma2013vae,rezende2014vae}:

\eq{
\mathcal{L}(\theta) \doteq \sum_{t=1}^T \Big(
&\beta\KL[\reprW(s_t|s_{t-1},a_{t-1},x_t) || \dynW(s_t|s_{t-1},a_{t-1})] \\[-1.5ex]
& + \|\recW(s_t)-x_t\|^2
  + (\rewW(s_{t+1})-r_t)^2
\Big)
\quad\text{where}\quad s_{1:T}\sim\mathrm{repr_\theta}
\label{eq:wmloss}
}

The variational objective encourages learning a Markovian sequence of latent states with the following properties: The states should be informative of the corresponding observations and rewards, the dynamics model should predict future states accurately, and the representations should be formed such that they are easy to predict.
The hyperparameter $\beta$ trades off the predictability of the latent states with the reconstruction quality \citep{beattie2016dmlab,alemi2018broken}.

\subsection{Goal Autoencoder}
\label{sec:ae}

The world model representations $s_t$ are 1024-dimensional continuous vectors.
Selecting such representations as goals would be challenging for the manager because this constitutes a very high-dimensional continuous action space.
To avoid a high-dimensional continuous control problem for the manager, Director compresses the representations $s_t$ into smaller discrete codes $z$ using a goal autoencoder that is trained on replay buffer model states from \cref{eq:wmloss}:

\eq{
\text{Goal Encoder:} \quad \encW(z|s_t) \qquad
\text{Goal Decoder:} \quad \decW(z) \approx s_t
}

Simply representing each model state $s_t$ by a class in one large categorical vector would require roughly one category per distinct state in the environment. 
It would also prevent the manager from generalizing between its different outputs. 
Therefore, we opt for a factorized representation of multiple categoricals.
Specifically, we choose the vector of categoricals approach introduced in DreamerV2 \cite{hafner2020dreamerv2}. 
As visualized in \cref{fig:vectorcat}, the goal encoder takes a model state as input and predicts a matrix of 8$\times$8 logits, samples a one-hot vector from each row, and flattens the results into a sparse vector with 8 out of 64 dimensions set to 1 and the others to 0. 
Gradients are backpropagated through the sampling by straight-through estimation \citep{bengio2013straight}. 
The goal autoencoder is optimized end-to-end by gradient descent on the variational objective:

\eq{
\mathcal{L}(\phi) \doteq
\big\|\, \decW(z) - s_t \,\big\|^2 + \beta\KL[\encW(z|s_t) || \p(z)]
\quad\text{where}\quad z \sim \encW(z|s_t)
\label{eq:aeloss}
}

The first term is a mean squared error that encourages the encoder to compute informative codes from which the input can be reconstructed.
The second term encourages the encoder to use all available codes by regularizing the distribution towards a uniform prior $\p(z)$.
The autoencoder is trained at the same time as the world model but does not contribute gradients to the world model.

\subsection{Manager Policy}
\label{sec:manager}

Director learns a manager policy that selects a new goal for the worker every fixed number of $K=8$ time steps.
The manager is free to choose goals that are much further than 8 steps away from the current state, and in practice, it often learns to choose the most distant goals that the worker is able to achieve. 
Instead of selecting goals in the high-dimensional continuous latent space of the world model, the manager outputs abstract actions in the discrete code space of the goal autoencoder (\Cref{sec:ae}). 
The manager actions are then decoded into world model representations before they are passed on to the worker as goals. 
To select actions in the code space, the manager outputs a vector of categorical distributions, analogous to the goal encoder in \cref{sec:ae}:

\eq{
\text{Manager Policy:} \quad \mgrW(z|s_t)
}

The objective for the manager is to maximize the discounted sum of future task rewards and exploration rewards.
The exploration encourages the manager to choose novel goals for the worker, resulting in temporally-abstract exploration.
This is important because the worker is goal-conditioned, so without an explicit drive to expand the state distribution, it could prefer going back to previously common states it has been trained on the most, and thus hinder exploration of new states in the environment.
Because the goal autoencoder is trained from the replay buffer, it tracks the current state distribution of the agent and we can reward novel states as those that have a high reconstruction error under the goal autoencoder:

\eq{
r^{\mathrm{expl}}_t \doteq \big\|\, \decW(z) - s_{t+1} \,\big\|^2\ \quad\text{where}\quad z \sim \encW(z|s_{t+1})
\label{eq:explrew}
}

Both manager and worker policies are trained from the same imagined rollouts and optimized using Dreamer \citep{hafner2019dreamer,hafner2020dreamerv2}, which we summarize in \cref{sec:policyopt}.
The manager learns two state-value critics for the extrinsic and exploration rewards, respectively.
The critics are used to fill in rewards beyond the imagination horizon and as baseline for variance reduction \citep{williams1992reinforce}.
We normalize the extrinsic and exploration returns by their exponential moving standard deviations before summing them with weights $w^{\mathrm{extr}}=1.0$ and $w^{\mathrm{expl}}=0.1$.
For updating the manager, the imagined trajectory is temporally abstracted by selecting every $K$-th model state and by summing rewards within each non-overlapping subsequence of length $K$.
No off-policy correction \citep{schulman2017ppo,nachum2018hiro} is needed because the imagined rollouts are on-policy.

\subsection{Worker Policy}

The worker is responsible for reaching the goals chosen by the manager.
Because the manager outputs codes $z$ in the discrete space of the goal autoencoder, we first decode the goals into the state space of the world model $g \doteq \dec(z)$.
Conditioning the worker on decoded goals rather than the discrete codes has the benefit that its learning becomes approximately decoupled from the goal autoencoder.
The worker policy is conditioned on the current state and goal, which changes every $K=8$ time steps, and it produces primitive actions $a_t$ to reach the feature space goal:

\eq{
\text{Worker Policy:} \quad \wkrW(a_t|s_t,g)
}

To reach its latent goals, we need to choose a reward function for the worker that measures the similarity between the current state $s_t$ and the current goal $g$, both of which are 1024-dimensional continuous activation vectors.
Natural choices would be the negative L2 distance or the cosine similarity and their choice depends on the underlying feature space, which is difficult to reason about.
Empirically, we found the cosine similarity to perform better.
Cosine similarity would usually normalize both vectors, but normalizing the state vector encourages the worker to remain near the origin of the latent space, so that it can quickly achieve any goal by moving a small amount in the right direction.
We thus incorporate the idea that the magnitude of both vectors should be similar into the cosine similarity, resulting in our \emph{max-cosine} reward:

\eq{
r^{\mathrm{goal}}_t \doteq \big(g/m\big)^T\big(s_{t+1}/m\big)
\quad\text{where}\quad m \doteq \max\!\big(\,\|g\|, \,\|s_{t+1}\|\,\big)
\label{eq:goalrew}
}

When the state and goal vectors are of the same magnitude, the reward simplifies to the cosine similarity.
When their magnitudes differ, the vectors are both normalized by the larger of the two magnitudes, and thus the worker receives a down-scaled cosine similarity as the reward.
As a result, the worker is incentivized to match the angle and magnitude of the goal.
Unlike the L2 similarity, the reward scale of our goal reward is not affected by the scale of the underlying feature space.

The worker maximizes only the goal rewards.
We make this design choice to demonstrate that the interplay between the manager and the worker is successful across many environments, although we also include an ablation experiment where the worker additionally receives task reward, which further improves performance.
The worker is optimized by Dreamer with a goal-conditioned state-value critic.
For updating the worker, we cut the imagined rollouts into distinct trajectories of length $K$ within which the goal is constant.
The state-critic estimates goal rewards beyond this horizon under the same goal, allowing the worker to learn to reach far-away goals.

\begin{figure}[t]
\vspace*{-5ex}
\centering
\includegraphics[width=\textwidth]{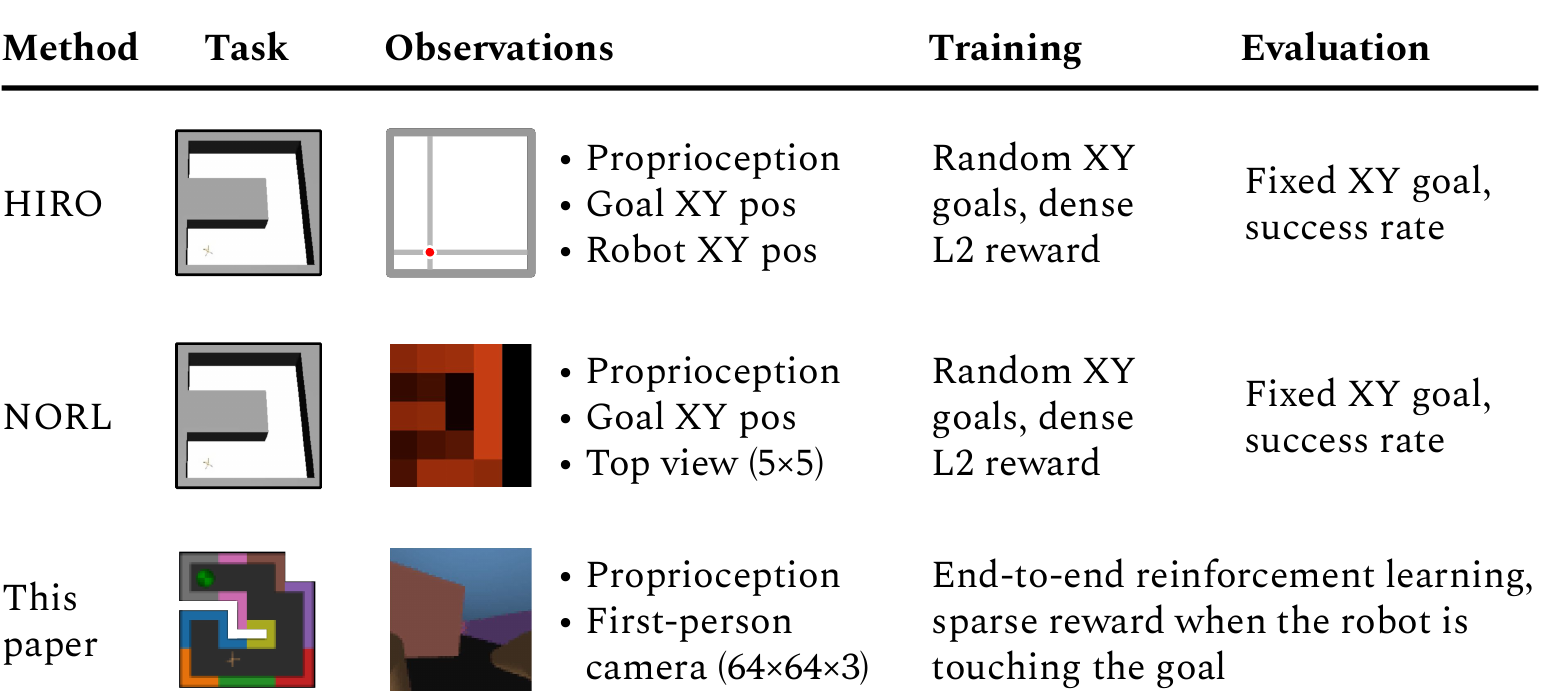}
\caption{
Comparison of Ant Mazes in the literature and this paper. 
HIRO \citep{nachum2018hiro} provided global XY coordinates of the goal and robot position to the agent and trained with dense rewards on uniformly sampled training goals. 
NORL \citep{nachum2018norl} replaced the robot XY position with a global top-down view of the environments, downsampled to 5$\times$5 pixels. 
In this paper, we tackle the more challenging problem of learning directly from an egocentric camera without global information provided to the agent, and only give a sparse reward for time steps where the robot is at the goal. 
To succeed at these tasks, an agent has to autonomously explore the environment and identify landmarks to localize itself and navigate the mazes.
}
\label{fig:inputs}
\end{figure}
\begin{figure}[t]
\vspace*{-5ex}
\centering
\hspace*{1.8em}
\includegraphics[width=0.93\textwidth]{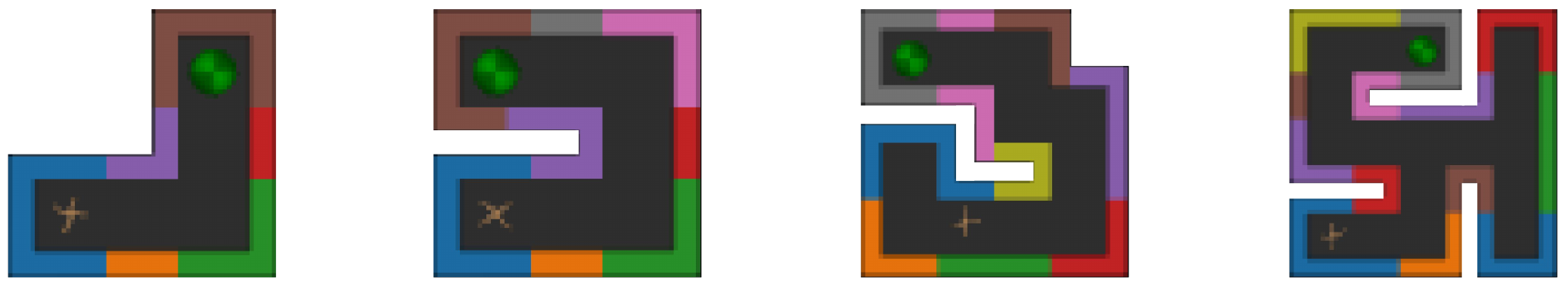} \\[2ex]
\includegraphics[width=\textwidth]{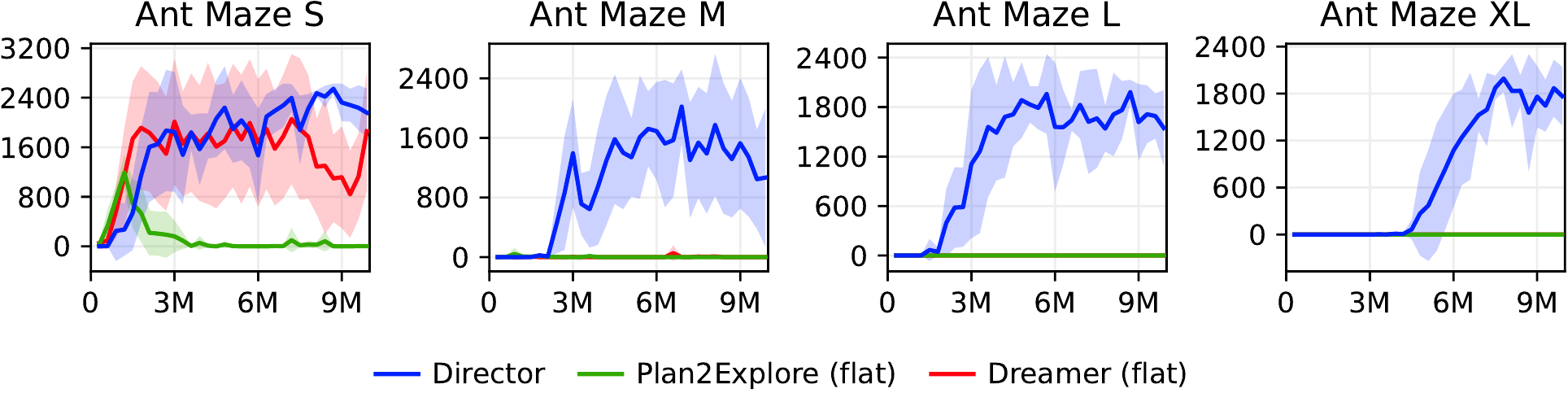}
\caption{Egocentric Ant Maze benchmark. A quadruped robot is controlled through joint torques to navigate to a fixed location in a 3D maze, given only first-person camera and proprioceptive inputs. This is in contrast to prior benchmarks where the agents received their global XY coordinate or top-down view. The only reward is given at time steps where the agent touches the reward object. Plan2Explore fails in the small maze because the robot flips over too much, a common limitation of low-level exploration methods. Director solves all four tasks by breaking them down into manageable subgoals that the worker can reach, while learning in the end-to-end reinforcement learning setting.}
\label{fig:mazes}
\end{figure}

\begin{figure}[t]
\vspace*{-3ex}
\centering
\hspace*{1em}
\includegraphics[width=0.942\textwidth]{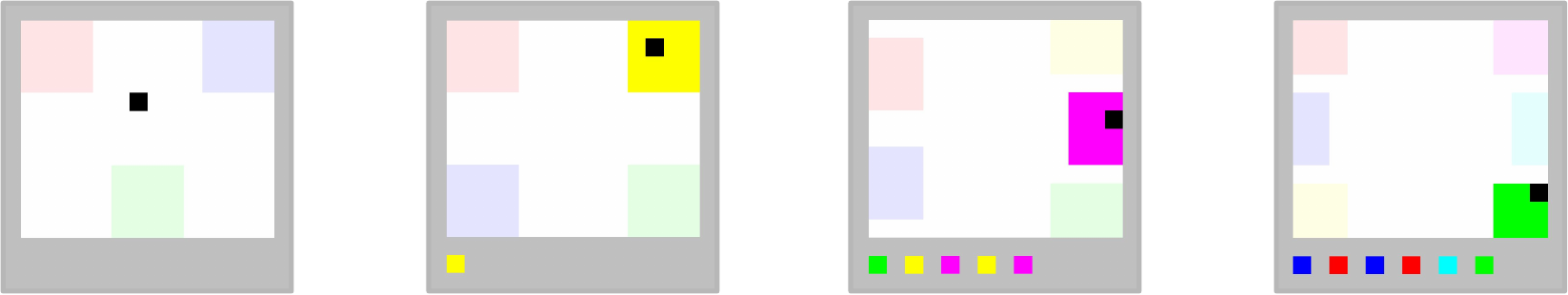} \\[2ex]
\includegraphics[width=\textwidth]{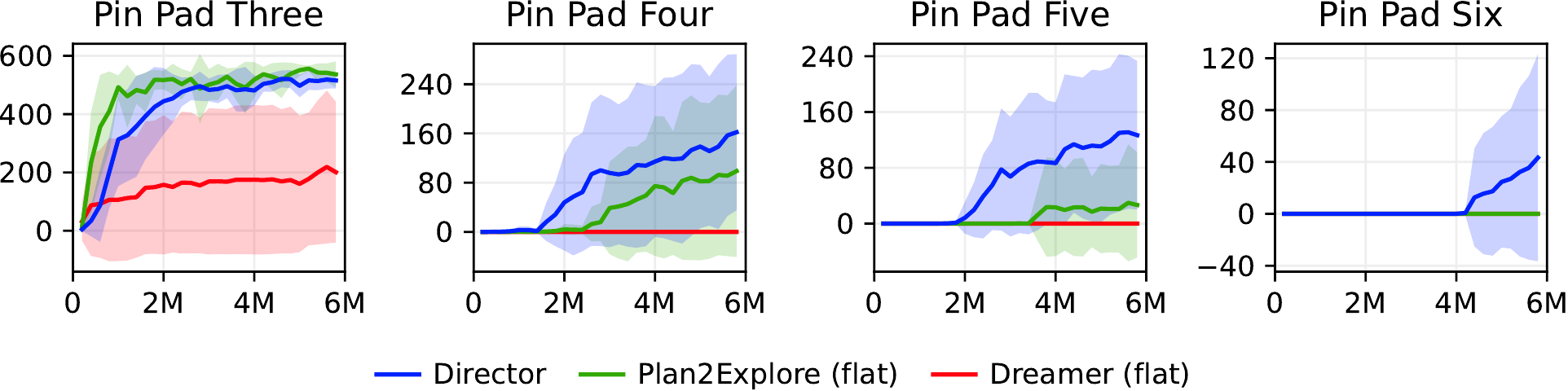}
\caption{%
Visual Pin Pad benchmark. 
The agent controls the black square to move in four directions. 
Each environment has a different number of pads that can be activated by walking to and stepping on them. 
A single sparse reward is given when the agent activates all pads in the correct sequence. 
The history of previously activated pads is shown at the bottom of the screen. 
Plan2Explore uses low-level exploration and performs well in this environment, but struggles for five and six pads, which requires more abstract exploration and longer credit assignment. 
Director learns successfully across all these environments, demonstrating its benefit on this long-horizon benchmark.
}
\label{fig:pinpads}
\end{figure}
\begin{figure}[t]
\vspace*{-3ex}
\centering
\includegraphics[width=\textwidth]{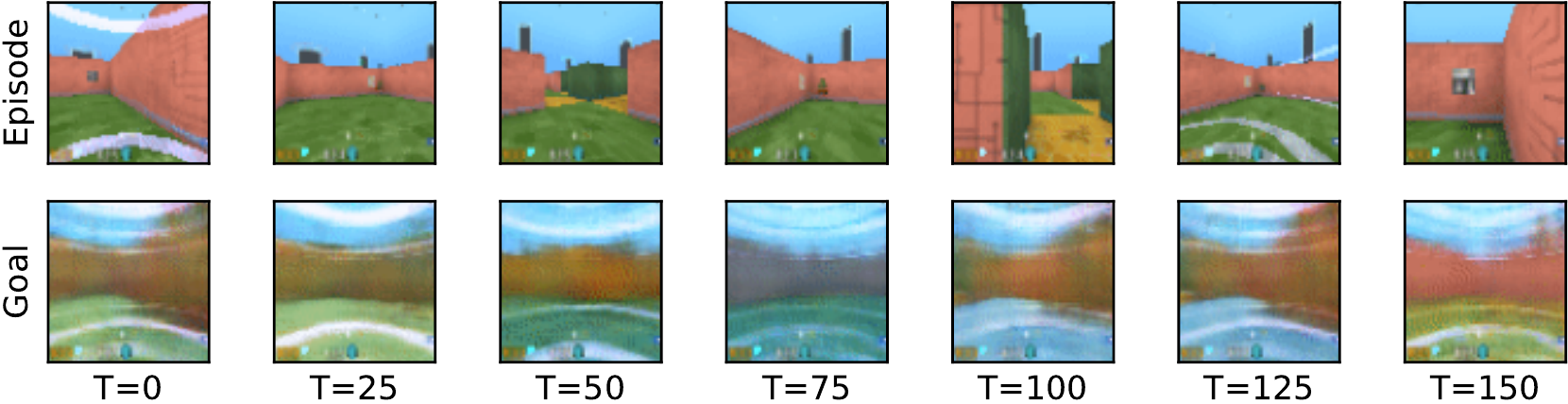} \\[2ex]
\includegraphics[width=\textwidth]{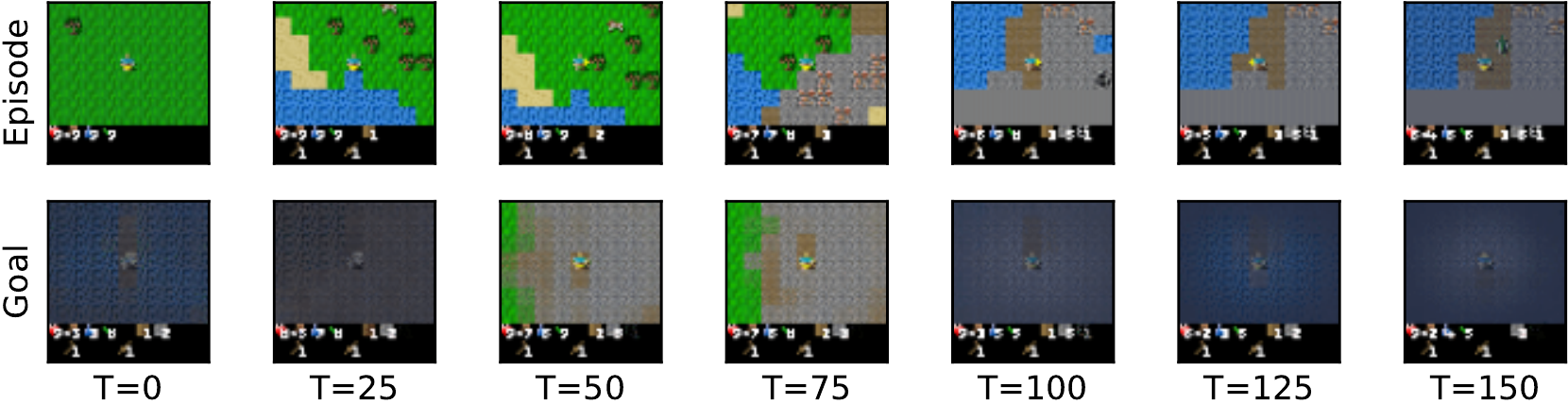} \\[2ex]
\includegraphics[width=\textwidth]{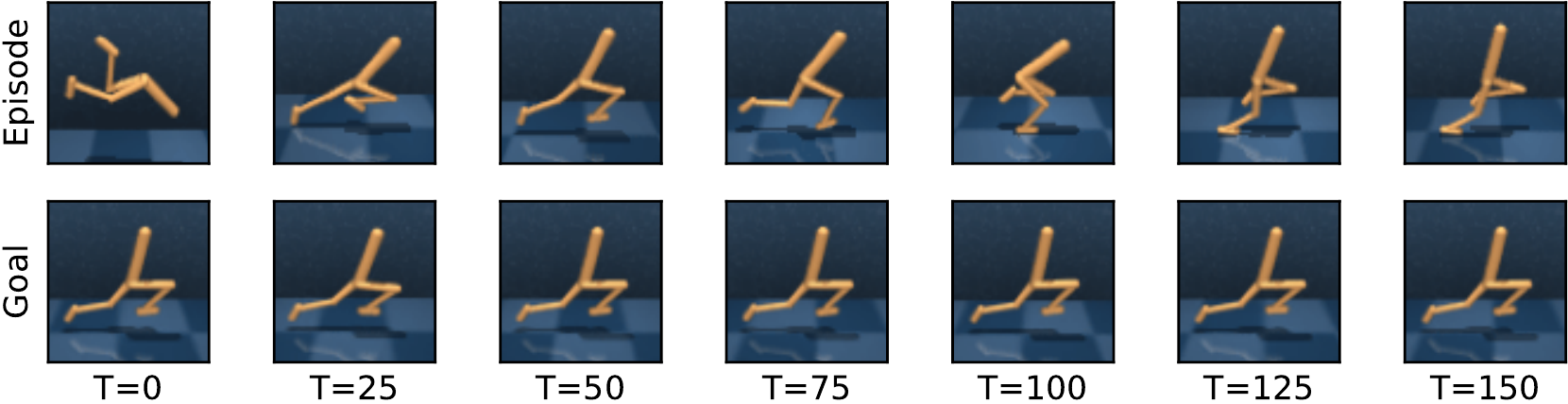}
\caption{
Subgoals discovered by Director. 
For interpretability, we decode the latent goals into images using the world model. 
\textbf{Top:} In DMLab, the manager chooses goals of the teleport animation that occurs every time the agent has collected the reward object, which can be seen in the episode at time step 125. 
\textbf{Middle:} In Crafter, the manager first directs the worker via the inventory display to collect wood and craft a pickaxe and a sword, with the worker following command.
The worker then suggests a cave via the terrain image to help the worker find stone.
As night breaks, it suggests hiding from the monsters in a cave or on a small island. 
\textbf{Bottom:} In Walker, the manager abstracts away the detailed leg movements by suggesting a forward leaning pose and a shifting floor pattern, with the worker successfully filling in the joint movements. Fine-grained subgoals are not required for this task, because the horizon needed for walking is short enough for the worker.
Videos are available on the project website: \url{https://danijar.com/director/}
}
\label{fig:selection}
\end{figure}

\section{Experiments}
\label{sec:experiments}

We evaluate Director on two challenging benchmark suites with visual inputs and very sparse rewards, which we expect to be challenging to solve using a flat policy without hierarchy (\Cref{sec:sparse}).
We further evaluate Director on a wide range of standard tasks from the literature to demonstrate its generality and ensure that the hierarchy is not harmful in simple settings (\Cref{sec:standard}). 
We use a fixed set of hyperparameters not only across tasks but also across domains, detailed in \cref{tab:hparams}. 
Finally, we offer insights into the learned hierarchical behaviors by visualizing the latent goals selected during environment episodes (\Cref{sec:interp}). 
Ablations and additional results are included in the appendix.

\paragraph{Implementation}

We implemented Director on top of the public source code of DreamerV2 \citep{hafner2020dreamerv2}, reusing its default hyperparameters. 
We additionally increased the number of environment instances to 4 and set the training frequency to one gradient step per 16 policy steps, which drastically reduced wall-clock time and decreased sample-efficiency mildly. 
Implementing Director in the code base amounts to about 250 lines of code. 
The computation time of Director is 20\% longer than that of DreamerV2. 
Each training run used a single V100 GPU with XLA and mixed precision enabled and completed in less than 24 hours. 
All our agents and environments will be open sourced upon publication to facilitate future research in hierarchical reinforcement learning.

\paragraph{Baselines}

To fairly compare Director to the performance of non-hierarchical methods, we compare to the DreamerV2 agent on all tasks. 
DreamerV2 has demonstrated strong performance on Atari \citep{hafner2020dreamerv2}, Crafter \citep{hafner2021crafter} and continuous control tasks \citep{yarats2021drqv2} and outperforms top model-free methods in these domains. 
We kept the default hyperparameters that the authors tuned for DreamerV2 and did not change them for Director. 
In addition to its hierarchical policy, Director employs an exploration bonus at the top-level. 
To isolate the effects of hierarchy and exploration, we compare to Plan2Explore \citep{sekar2020plan2explore}, which uses ensemble disagreement of forward models as a directed exploration signal. 
We combined the extrinsic and exploration returns after normalizing by their exponential moving standard deviation with weights $1.0$ and $0.1$, as in Director. 
We found Plan2Explore to be effective across both continuous and discrete control tasks.

\subsection{Sparse Reward Benchmarks}
\label{sec:sparse}

\paragraph{Egocentric Ant Mazes}

Learning navigation tasks directly from joint-level control has been a long-standing milestone for reinforcement learning with sparse rewards, commonly studied with quadruped robots in maze environments \citep{florensa2017snn,nachum2018hiro}. 
However, previous attempts typically required domain-specific inductive biases to solve such tasks, such as providing global XY coordinates to the agent, easier practice goals, and a ground-truth distance reward, as summarized in \cref{fig:inputs}. 
In this paper, we instead attempt learning directly from first-person camera inputs, without privileged information, and a single sparse reward that the agent receives while in the fixed goal zone. 
The control frequency is 50Hz and episodes end after a time limit of 3000 steps. 
There are no early terminations that could leak task-information to the agent \citep{laskin2022cic}. 
To help the agent localize itself from first-person observations, we assign different colors to different walls of the maze.

As shown in \cref{fig:mazes}, we evaluate the agents in four mazes that span varying levels of difficulty. 
Because of the sparse reward, the episode returns correspond to the number of time steps for which the agent remains at the goal after reaching it. 
Curves show the mean and standard deviation across 5 independent seeds. 
We find that the flat Dreamer agent succeeds at the smallest of the four mazes, and the flat exploration policy Plan2Explore makes some initial learning progress but fails to converge to the optimal solution. 
Inspecting the trajectories revealed that Plan2Explore chooses too chaotic actions that often result in the robot flipping over. 
None of the baselines learn successful behaviors in the larger mazes, demonstrating that the benchmark pushes the limits of current approaches. 
In contrast, Director discovers and learns to reliably reach the goal zone at all four difficulty levels, with the larger maze taking longer to master.

\paragraph{Visual Pin Pads}

The Pin Pad suite of environments is designed to evaluate an agent's ability to explore and assign credit over long horizons, isolated from the complexity of 3D observations or sophisticated joint-level control. 
As shown in \cref{fig:pinpads}, each environment features a moving black square that the agent can control in all four directions and fixed pads of different colors that the agent can activate by walking over to the pad and stepping on it. 
The task requires discovering the correct sequence of activating all pads, at which point the agent receives a sparse reward of 10 points and the agent position is randomly reset. 
Episodes last for 2000 steps and there are no intermediate rewards for activating the pads. 
To remove the orthogonal challenge of learning long-term memory \citep{gregor2019simcore}, the history of previously activated pads is displayed at the bottom of the image.

The agent performance is shown in \cref{fig:pinpads}, which displays mean and standard deviation across five independent seeds. 
The easiest environment contains three pads, so the agent only has to decide whether to activate the pads in clockwise or counter-clockwise sequence. 
The flat Dreamer agent sometimes discovers the correct sequence. 
The exploration bonus of Plan2Explore offers a significant improvement over Dreamer on this task. 
Dreamer fails to discover the correct sequence in the harder environments that contain more pads. 
Plan2Explore still achieves some reward with four pads, struggles with five pads, and completely fails with six pads. 
In contrast, Director discovers the correct sequence in all four environments, demonstrating the benefit of hierarchy over flat exploration.

\subsection{Standard Benchmarks}
\label{sec:standard}

To evaluate the robustness of Director, we train on a wide range of standard benchmarks, which typically require no long-term reasoning. 
We choose Atari games \citep{bellemare2013ale}, the Control Suite from pixels \citep{tassa2018dmcontrol}, Crafter \citep{hafner2021crafter}, and tasks from DMLab \citep{beattie2016dmlab} to cover a spectrum of challenges, including continuous and discrete actions and 2D and 3D environments. 
We compare two versions of Director. 
In its standard variant, the worker learns purely from goal rewards. 
This tests the ability of the manager to propose successful goals and the ability of the worker to achieve them. 
In the second variant, the worker learns from goal and task returns with weights $1.0$ and $0.5$, allowing the worker to fill in low-level details in a task-specific manner, which the manager may be too coarse-grained to provide. 
Ideally, we would like to see Director not perform worse than Dreamer on any task when giving task reward to the worker.

The results of this experiment are summarized in \cref{sec:standardfig} due to space constraints, with the full training curves for Atari and the Control Suite included in \cref{sec:atari,sec:dmc}.
We observe that Director indeed learns successfully across many environments, showing broader applicability than most prior hierarchical reinforcement learning methods. 
In addition, providing task reward to the worker is not as important as expected --- the hierarchy solves a wide range of tasks purely by following goals at the low level. 
Additionally providing task reward to the worker completely closes the gap to the state-of-the-art DreamerV2 agent.
\Cref{fig:atari} in the appendix further shows that Director achieves a higher human-normalized median score than Dreamer on the 55 Atari games.

\vspace*{3ex}
\pagebreak

\subsection{Goal Interpretations}
\label{sec:interp}

To gain insights into the decision making of Director, we visualize the sequences of goals it selects during environment interaction. 
While the goals are latent vectors, the world model allows us to decode them into images for human inspection. 
\Cref{fig:first,fig:selection} show episodes with the environment frames (what the agent sees) at the top and the decoded subgoals (what the manager wants the worker to achieve) at the bottom. 
Visualizations for additional environments are included in \cref{sec:moregoals}.

\begin{itemize}

\itempar{Ant Maze M}
The manager chooses goals that direct the agent through the different sections of the maze until the reward object is reached. 
We also observed that initially, the manager chooses more intermediate subgoals whereas later during training, the worker learns to achieve further away goals and thus the manager can select fewer intermediate goals.

\itempar{DMLab Goals Small}
The manager directs the worker to the teleport animation, which occurs every time the agent collects the reward object and gets teleported to a random location in the 3D maze.
Time step $T=125$ shows an example of the worker reaching the animation in the environment. 
Unlike the Ant Maze benchmark, navigating to the goal in DMLab requires no joint-level control and is simple enough for the worker to achieve without fine-grained subgoals.


\itempar{Crafter}
The manager requests higher inventory counts of wood and stone materials, as well as wooden tools, via the inventory display at the bottom of the screen. 
It also directs the worker to a cave to find stone and coal but generally spends less effort on suggesting what the world around the agent should look like.
As night breaks, the manager tasks the worker with finding a small cave or island to hide from the monsters that are about to spawn.

\itempar{Walker Walk}
The manager abstracts away the detail of leg movement, steering the worker using a forward leaning pose with both feet off the ground and a shifting floor pattern. 
While the images cannot show this, the underlying goal vectors are Markovian states that can contain velocities, so it is likely that the manager additionally requests high forward velocity.
The worker fills in the details of standing up and moving the legs to pass through the sequence of goals.

\end{itemize}
\vspace*{-.5ex}

Across domains, Director chooses semantically meaningful goals that are appropriate for the task, despite using the same hyperparameters across all tasks and receiving no domain-specific knowledge.

\vspace*{-3ex}
\section{Related Work}
\label{sec:related}
\vspace*{-1.5ex}

Approaches to hierarchical reinforcement learning include learning low-level policies on a collection of easier pre-training tasks \citep{heess2016modulated,tessler2017dsn,frans2017mlsh,rao2021helms,veeriah2021modac}, discovering latent skills via mutual-information maximization \citep{gregor2016vic,florensa2017snn,hausman2018skills,eysenbach2018diayn,achiam2018valor,merel2018humanoid,laskin2022cic,sharma2019dads,xie2020lsp,hafner2020apd,strouse2021disdain}, or training the low-level as a goal-conditioned policy \citep{andrychowicz2017her,levy2017hac,nachum2018hiro,nachum2018norl,co2018sectar,warde2018discern,nair2018rig,pong2019skewfit,hartikainen2019ddl,gehring2021hsd3,shah2021recon}. 
These approaches are described in more detail in \cref{sec:morerelated}.

Relatively few works have demonstrated successful learning of hierarchical behaviors directly from pixels without domain-specific knowledge, such as global XY positions, manually specified pre-training tasks, or precollected diverse experience datasets.
HSD-3 \citep{gehring2021hsd3} showed transfer benefits for low-dimensional control tasks. 
HAC \citep{levy2017hac} learned interpretable hierarchies but required semantic goal spaces. 
FuN \citep{vezhnevets2017fun} learned a two-level policy where both levels maximize task reward and the lower level is regularized by a goal reward but did not demonstrate clear benefits over an LSTM baseline \citep{hochreiter1997lstm}. 
We leverage explicit representation learning and temporally-abstract exploration, demonstrate substantial benefits over flat policies on sparse reward tasks, and underline the generality of our method by showing successful learning without giving task reward to the worker across many domains.

\vspace*{-1ex}
\section{Discussion}
\label{sec:discussion}
\vspace*{-1.5ex}

We present Director, a reinforcement learning agent that learns hierarchical behaviors from pixels by planning in the latent space of a learned world model. 
To simplify the control problem for the manager, we compress goal representations into compact discrete codes. 
Our experiments demonstrate the effectiveness of Director on two benchmark suites with very sparse rewards from pixels. 
We also show that Director learns successfully across a wide range of different domains without giving task reward to the worker, underlining the generality of the approach. 
Decoding the latent goals into images using the world model makes the decision making of Director more interpretable and we observe it learning a diverse range of strategies for breaking tasks down into subgoals.

\paragraph{Limitations and future work}

We designed Director to be as simple as possible, at the expense of design choices that could restrict its performance.
The manager treats its action space as a black box, without any knowledge that its actions correspond to states. For example, one could imagine regularizing the manager to choose goals that have a high value under its learned critic. 
Lifting the assumption of changing goals every fixed number of time steps, for example by switching based on a separate classifier or as soon as the previous goal has been reached, could enable the hierarchy to adapt to the environment and perform better on tasks that require precise timing. 
Moreover, goals are points in latent space, whereas distributional goals or masks would allow the manager to only specify targets for the parts of the state that are currently relevant.
Learning temporally-abstract dynamics would allow efficiently learning hierarchies of more than two levels without having to use exponentially longer batches.
We suspect that these ideas will improve the long-term reasoning of the agent.
Besides improving the capabilities of the agent, we see disentangling the reasons for why Director works well, beyond the ablation experiments in the appendix, as promising future work.

\paragraph{Acknowledgements}

We thank Volodymyr Mnih, Michael Laskin, Alejandro Escontrela, Nick Rhinehart, and Hao Liu for insightful discussions. We thank Kevin Murphy, Ademi Adeniji, Olivia Watkins, Paula Gradu, and Younggyo Seo for feedback on the initial draft of the paper.

\begin{hyphenrules}{nohyphenation}
\bibliography{references}
\end{hyphenrules}
\clearpage
\section*{Societal Impact}

Developing hierarchical reinforcement learning methods that are useful for real-world applications will still require further research.
However, in the longer term future, it has the potential to help humans automate more tasks by reducing the need to specify intermediate rewards. 
Learning autonomously from sparse rewards has the benefit of reduced human effort and less potential for reward hacking, but reward hacking is nonetheless a possibility that will need to be addressed once hierarchical reinforcement learning systems become more powerful and deployed around humans.

As reinforcement learning techniques become deployed in real-world environments, they will be capable of causing direct harm, intentional or unintentional.
Director does not attempt to directly address such safety issues, but the ability to decode its latent goals into images for human inspection provides some transparency to the decisions the agent is making.
This transparency may permit auditing of failures after-the-fact, and may additionally support interventions in some situations, if a human or automated observer is in a position to monitor the high-level goals Director generates.

\section*{Checklist}

\begin{enumerate}

\item For all authors...
\begin{enumerate}
  \item Do the main claims made in the abstract and introduction accurately reflect the paper's contributions and scope?
    \answerYes{}
  \item Did you describe the limitations of your work?
    \answerYes{} See \cref{sec:discussion}.
  \item Did you discuss any potential negative societal impacts of your work?
    \answerYes{}
  \item Have you read the ethics review guidelines and ensured that your paper conforms to them?
    \answerYes{}
\end{enumerate}

\item If you are including theoretical results...
\begin{enumerate}
  \item Did you state the full set of assumptions of all theoretical results?
    \answerNA{}
        \item Did you include complete proofs of all theoretical results?
    \answerNA{}
\end{enumerate}

\item If you ran experiments...
\begin{enumerate}
  \item Did you include the code, data, and instructions needed to reproduce the main experimental results (either in the supplemental material or as a URL)?
    \answerYes{} As stated in the text, we will provide code and training curves upon publication.
  \item Did you specify all the training details (e.g., data splits, hyperparameters, how they were chosen)?
    \answerYes{} See \cref{tab:hparams}.
        \item Did you report error bars (e.g., with respect to the random seed after running experiments multiple times)?
    \answerYes{} See \cref{fig:mazes,fig:pinpads,fig:standard,fig:dmc,fig:atari}.
        \item Did you include the total amount of compute and the type of resources used (e.g., type of GPUs, internal cluster, or cloud provider)?
    \answerYes{} See \cref{sec:experiments}.
\end{enumerate}

\item If you are using existing assets (e.g., code, data, models) or curating/releasing new assets...
\begin{enumerate}
  \item If your work uses existing assets, did you cite the creators?
    \answerNA{}
  \item Did you mention the license of the assets?
    \answerNA{}
  \item Did you include any new assets either in the supplemental material or as a URL?
    \answerNA{}
  \item Did you discuss whether and how consent was obtained from people whose data you're using/curating?
    \answerNA{}
  \item Did you discuss whether the data you are using/curating contains personally identifiable information or offensive content?
    \answerNA{}
\end{enumerate}

\item If you used crowdsourcing or conducted research with human subjects...
\begin{enumerate}
  \item Did you include the full text of instructions given to participants and screenshots, if applicable?
    \answerNA{}
  \item Did you describe any potential participant risks, with links to Institutional Review Board (IRB) approvals, if applicable?
    \answerNA{}
  \item Did you include the estimated hourly wage paid to participants and the total amount spent on participant compensation?
    \answerNA{}
\end{enumerate}

\end{enumerate}

\appendix
\counterwithin{figure}{section}
\counterwithin{table}{section}
\clearpage
\section{Standard Benchmarks}
\label{sec:standardfig}
\begin{figure}[t]
\centering
\vspace*{-4ex}
\includegraphics[width=\textwidth]{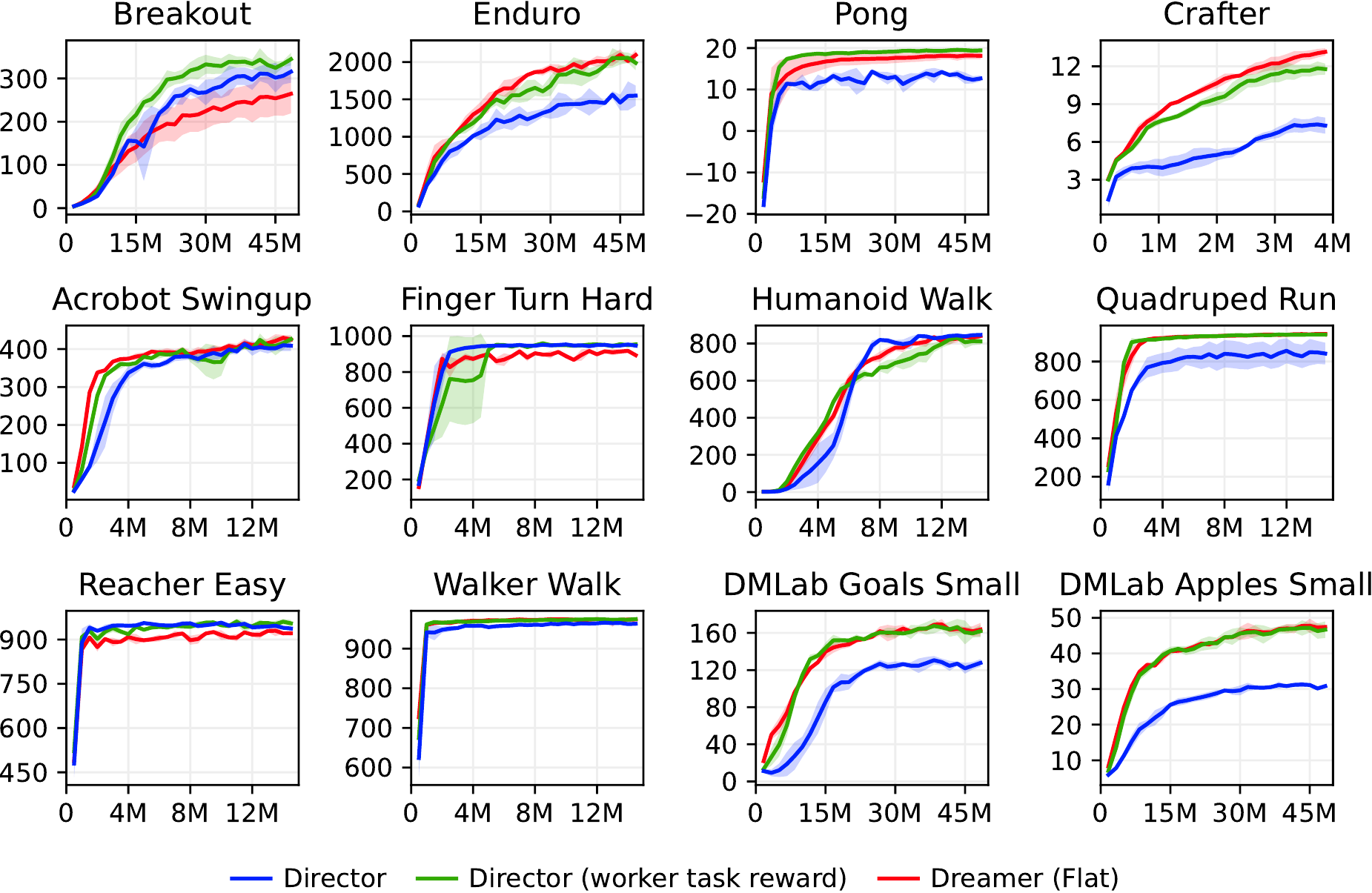}
\caption{%
Evaluation of Director on standard benchmarks, showing that Director is not limited to sparse reward tasks but is generally applicable. 
Director learns successfully across Atari, Crafter, Control Suite, and DMLab. 
This is an accomplishment because the worker receives no task reward and is purely steered through sub-goals selected by the manager. 
When additionally providing task reward to the worker, performance matches that of the state-of-the-art model-based agent Dreamer.
}
\label{fig:standard}
\end{figure}

\section{Ablation: Goal Autoencoder}
\begin{figure}[h!]
\centering
\includegraphics[width=\textwidth]{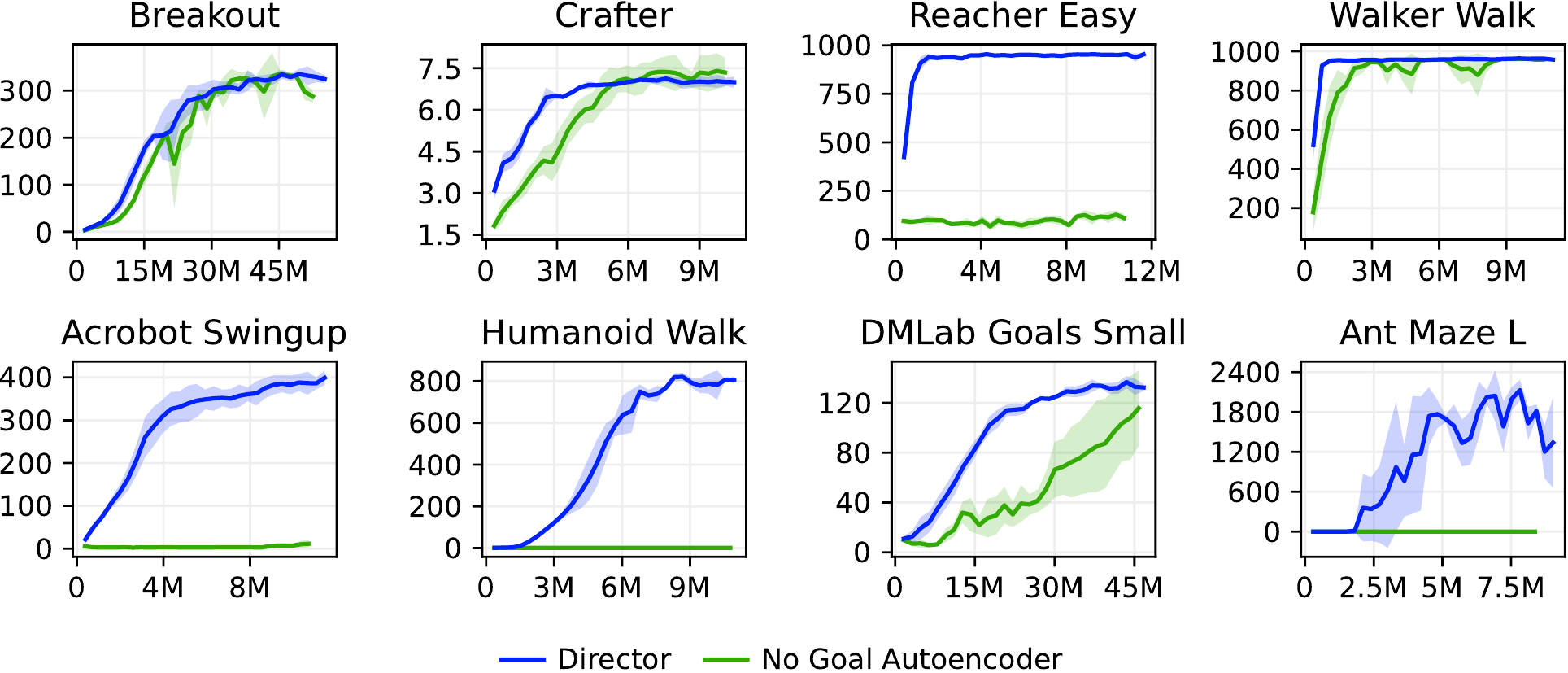}
\caption{
Ablation of the Goal Autoencoder used by Director (\Cref{sec:ae}). 
We compare the performance of Director to that of a hierarchical agent where the manager directly chooses 1024-dimensional goals in the continuous representation space of the world model. 
We observe that this simplified approach works surprisingly well in some environments but fails at environments with sparser rewards, likely because the control problem becomes too challenging for the manager.
}
\label{fig:ablation_ae}
\end{figure}

\clearpage
\section{Ablation: Goal Reward}
\begin{figure}[h!]
\centering
\includegraphics[width=\textwidth]{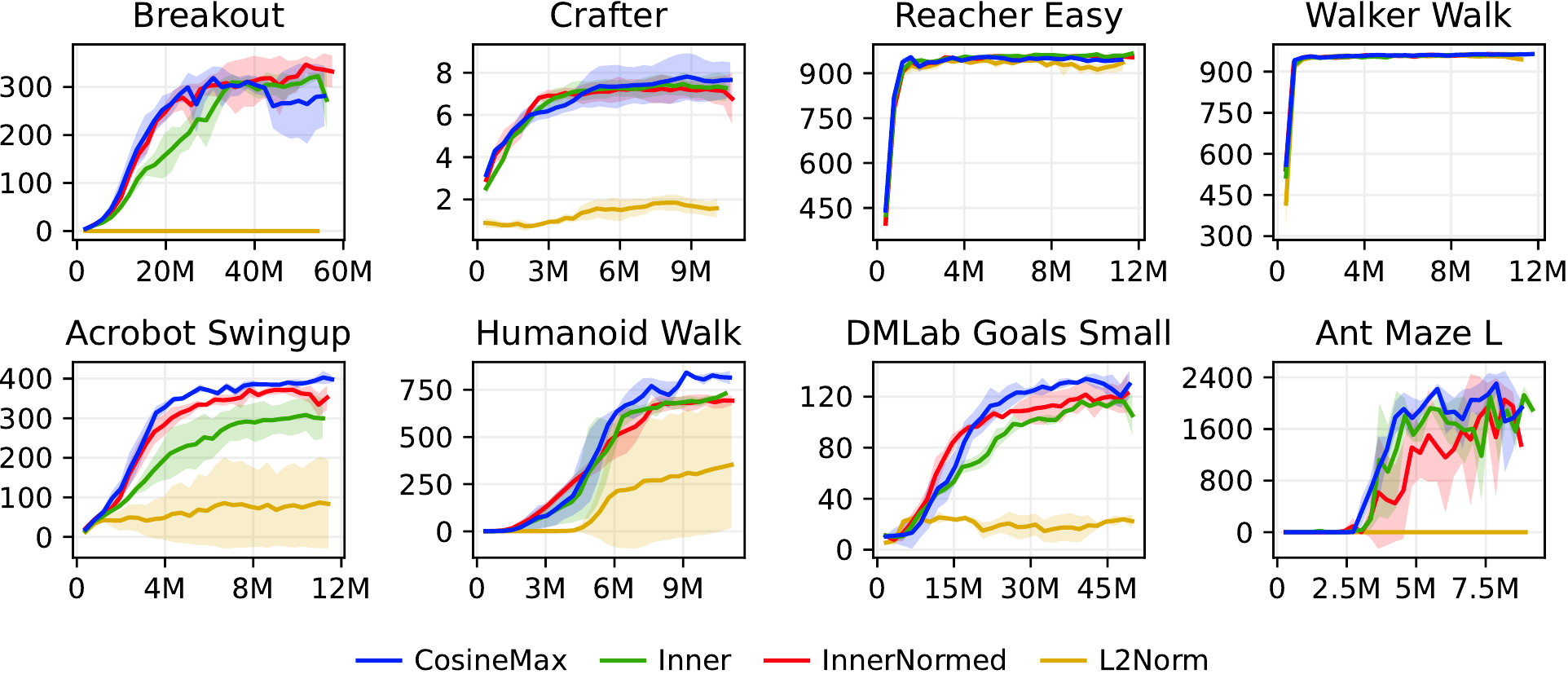}
\caption{
Ablation of the goal reward used by Director. 
We compare the performance of Director with cosine-max reward (\Cref{eq:goalrew}) to alternative goal similarity functions. 
\emph{Inner} refers to the inner product $g^T s_{t+1}$ without any normalization or clipping, which results in different reward scale based on the goal magnitude and encourages the worker to overshoot its goals in magnitude. 
\emph{InnerNormed} refers to $(g / \|g\|)^T (s_{t+1} / \|g\|)$ where the goal and state are normalized by the goal magnitude, which normalizes the reward scale across goals but still encourages the worker to overshoot its goals. 
\emph{L2Norm} is the negative euclidean distance $-\|g - s_{t+1}\|$. 
We observe that Director is robust to the precise goal reward, with the three rewards based on inner products performing well across tasks. 
The L2 reward works substantially worse. 
We hypothesize the reason to be that inner products allow goals to ignore some state dimensions by setting them to zero, whereas setting dimensions to zero for the L2 reward still requires the worker to care about them.
}
\label{fig:ablation_dist}
\end{figure}

\section{Ablation: Exploration}
\begin{figure}[h!]
\centering
\vspace*{-1ex}
\includegraphics[width=\textwidth]{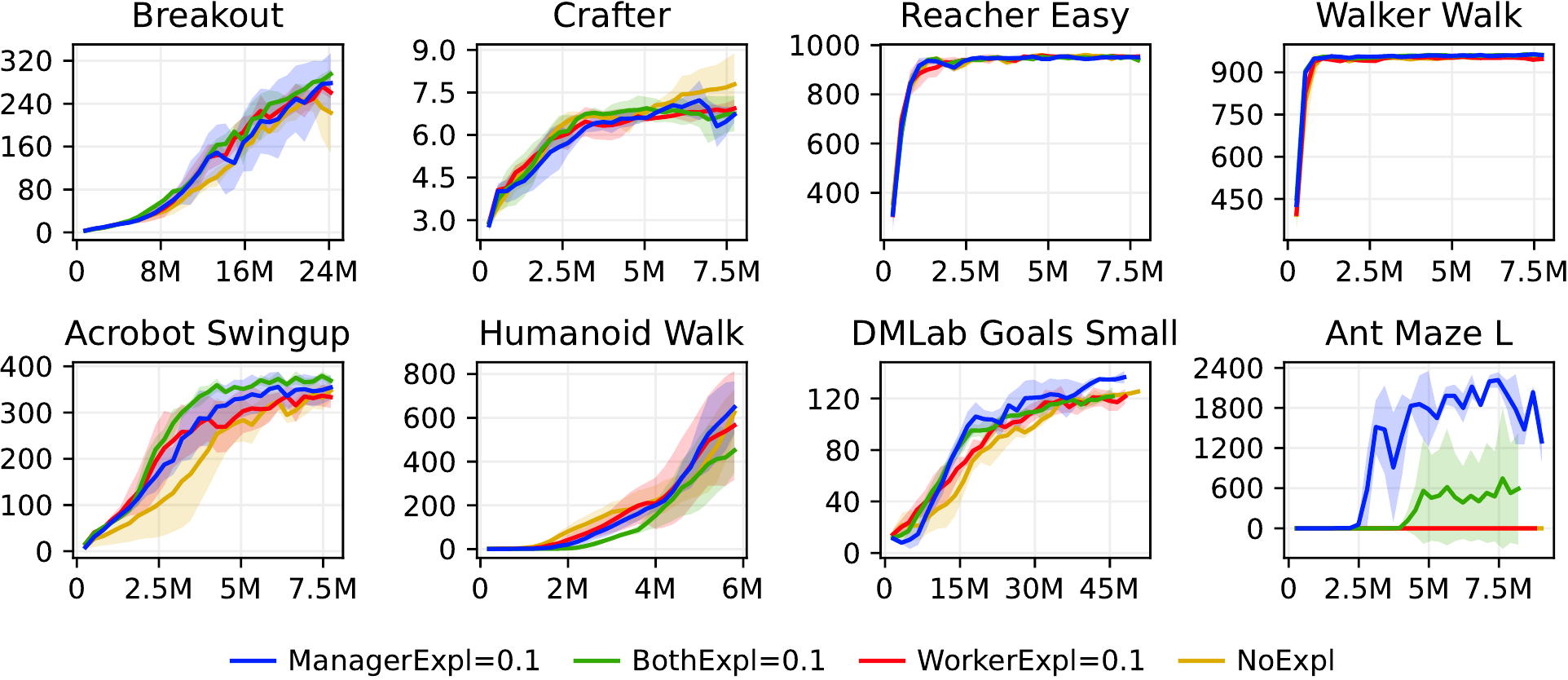}
\caption{
Ablation of where to apply the exploration bonus in Director (\Cref{eq:explrew}). 
We compare the performance of Director with exploration reward for only the manager, for only the worker, for both, and for neither of them. 
We find giving the exploration bonus to the manager, which results in temporally-abstract exploration, is required for successful learning in the Ant Maze, and that additional low-level exploration hurts because it results in too chaotic leg movements. 
In standard benchmarks that typically require short horizons, the exploration bonus is not needed. 
}
\label{fig:ablation_expl}
\end{figure}

\clearpage
\section{Pseudocode}
\label{sec:algo}
\begin{algorithm}[h]
\SetEndCharOfAlgoLine{}
\SetKwComment{Comment}{// }{}
Initialize replay buffer and neural networks. \;
\While{not converged}{

  \BlankLine\Comment{Acting}
  Update model state $s_t \sim \operatorname{repr}(s_t \mid s_{t-1},a_{t-1},x_t)$. \;
  
  \BlankLine
  \If{$t \operatorname{mod} K = 0$} {
    \Comment{Update internal goal}
    Sample abstract action $z\sim\mgr(z|s_t)$. \;
    Decode into model state $g=\dec(z)$. \;
  }
  \BlankLine
  
  Sample action $a_t\sim\wkr(a_t|s_t,g)$. \;
  Send action to environment and observe $r_t$ and $x_{t+1}$. \;
  Add transition $(x_t,a_t,r_t)$ to replay buffer. \;
  
  \BlankLine\Comment{Learning}
  \If{$t \operatorname{mod} E = 0$} {
  
    \BlankLine\Comment{World Model}
    Draw sequence batch $\{(x,a,r)\}$ from replay buffer. \;
    Update world model on batch (\Cref{eq:wmloss}) and get states $\{s\}$. \;
    
    \BlankLine\Comment{Goal Autoencoder}
    Update goal autoencoder on $\{s\}$ (\Cref{eq:aeloss}). \;
    
    \BlankLine\Comment{Policies}
    Imagine trajectory $\{(\hat{s},\hat{a},\hat{g},\hat{z})\}$ under model and policies starting from $\{s\}$. \;
    Predict extrinsic rewards $\{\operatorname{rew}(s)\}$. \;
    Compute exploration rewards $\smash{\{r^{\mathrm{expl}}\}}$ (\Cref{eq:explrew}). \; 
    Compute goal rewards $\smash{\{r^{\mathrm{goal}}\}}$ (\Cref{eq:goalrew}). \;
    Abstract trajectory to update manager (\Cref{eq:vloss,eq:aloss}). \;
    Split trajectory to update worker (\Cref{eq:vloss,eq:aloss}). \;
    
  }
}
\caption{Director}
\label{alg:agent}
\end{algorithm}

\section{Hyperparameters}
\label{sec:hparams}
\begin{table}[h!]
\centering
\begin{mytabular}{
  colspec = {| L{14em} | C{4em} | C{14em} |},
  row{1} = {font=\bfseries},
}

\toprule
\textbf{Name} & \textbf{Symbol} & \textbf{Value} \\
\midrule
Parallel envs & --- & $4$ \\
Training every & $E$ & $16$ \\
MLP size & --- & 4 $\times$ 512 \\
Activation & --- & $\operatorname{LayerNorm}+\operatorname{ELU}$ \\
Imagination horizon & $H$ & 16 \\
Discount factor & $\gamma$ & $0.99$ \\
Goal duration & $K$ & 8 \\
Goal autoencoder latents & $L$ & $8$ \\
Goal autoencoder classes & $C$ & $8$ \\
Goal autoencoder beta & $\beta$ & $1.0$ \\
Learning rate & --- & $10^{-4}$ \\
Weight decay & --- & $10^{-2}$ \\
Adam epsilon & $\epsilon$ & $10^{-6}$ \\
\bottomrule

\end{mytabular}
\vspace{1ex}
\caption{
Hyperparameters of Director. 
We use the same hyperparameters across all experiments. 
The hyperparameters for training the world model and optimizing the policies were left unchanged compared to DreamerV2 \citep{hafner2020dreamerv2}.
}
\label{tab:hparams}
\end{table}

\clearpage
\section{Vector of Categoricals}
\label{sec:veccat}
\begin{figure}[h]
\centering
\includegraphics[width=\textwidth]{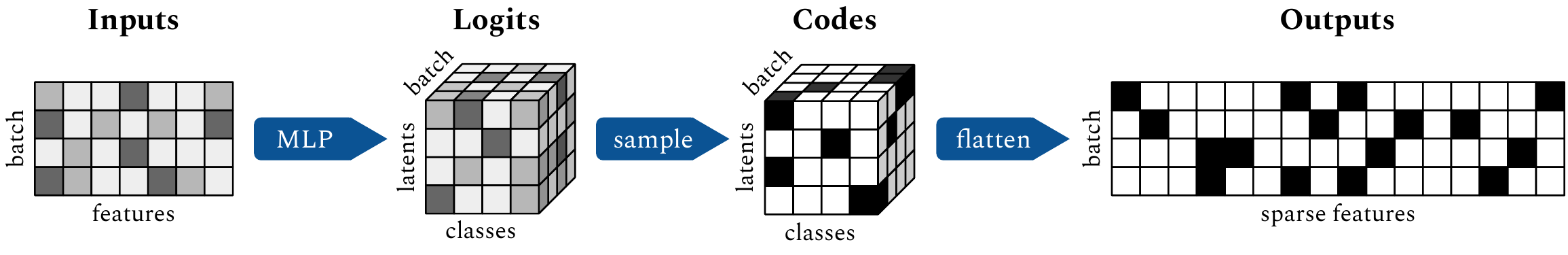}
\caption{
Director uses a variational autoencoder to turn its state representations into discrete tokens that are easy to select between for the manager policy. 
For this, we use the vector of categoricals approach of DreamerV2 \citep{hafner2020dreamerv2}. 
For a given input vector, the encoder outputs a matrix of logits of $L$ latent dimensions with $C$ classes each. 
We sample from the logits and one-hot encode the result to obtain a sparse matrix of the same shape as the logits. 
To backpropagate gradients through this step, we simply use the gradient with respect to the one-hot matrix as the gradient with respect to the categorical probabilities \citep{bengio2013straight}. 
The matrix is then flattened, resulting in a sparse representation with $L$ out of the $L \times C$ feature dimensions set to one.
}
\label{fig:vectorcat}
\end{figure}

\section{Policy Optimization}
\label{sec:policyopt}

Both the worker and manager policies of Director are optimized using the actor critic algorithm of Dreamer \citep{hafner2019dreamer,hafner2020dreamerv2}. 
For this, we use the world model to imagine a trajectory in its compact latent space. 
The actor network is optimized via policy gradients \citep{williams1992reinforce} with a learned state-value critic for variance reduction and to estimate rewards beyond the rollout:

\eq{
\text{Actor:} \quad \p<\pi>(a_t|s_t) \qquad
\text{Critic:} \quad v(s_t)
}

The world model allows cheaply generating as much on-policy experience as needed, so no importance weighting or clipping applies \citep{schulman2017ppo}. 
To train both actor and critic from an imagined trajectory of length $H$, $\lambda$-returns are computed from the sequence of rewards and predicted values \citep{sutton2018rlbook}:

\eq{
V^\lambda_t \doteq r_t + \gamma \Big(
  (1 - \lambda) v(s_{t+1}) +
  \lambda V^\lambda_{t+1}
\Big), \quad
V^\lambda_H \doteq v(s_H).
}

The $\lambda$-returns are averages over multi-step returns of different lengths, thus finding a trade-off between incorporating further ahead rewards quickly and reducing the variance of long sampled returns. 
The critic is learned by regressing the $\lambda$-returns via a squared loss, where $\operatorname{sg}(\cdot)$ indicates stopping the gradient around the targets:

\eq{
\mathcal{L}(v) \doteq
\E_{p_\phi,\pi}<\big>[ \textstyle\sum_{t=1}^{H-1}
  \frac{1}{2}\big(v(s_t)-\operatorname{sg}(V^\lambda_t)\big)^2
].
\label{eq:vloss}
}

The actor is updated by policy gradients on the same $\lambda$-returns, from which we subtract the state-value $v(s_t)$ as a baseline that does not depend on the current action for variance reduction. 
The second term in the actor objective, weighted by the scalar hyperparameter $\eta$, encourages policy entropy to avoid overconfidence and ensures that the actor explores different actions:

\eq{
\mathcal{L}(\pi)\doteq
-\E_{\pi,p_\phi}[\textstyle\sum_{t=1}^H
    \ln\p<\pi>(a_t|s_t)\operatorname{sg}(V^\lambda_t-v(s_t))
    +\,\eta\H[\p<\pi>(a_t|s_t)]
]
\label{eq:aloss}
}

When there are multiple reward signals, such as the task and exploration rewards for the manager, we learn separate critics \citep{burda2018rnd} for them and compute separate returns, which we normalize by their exponential moving standard deviation with decay rate $0.999$. 
The baselines are normalized by the same statistics and the weighted average of the advantages is used for updating the policy.

\clearpage
\section{Further Related Work}
\label{sec:morerelated}

\paragraph{Pretraining tasks}

One way to integrate domain knowledge into hierarchical agents is to learn primitives on simpler tasks and then compose them to solve more complex tasks afterwards. 
Learning primitives from manually specified tasks simplifies learning but requires human effort and limits the generality of the skills. 
DSN \citep{tessler2017dsn} explicitly specifies reward functions for the low-level policies and then trains a high-level policy on top to solve Minecraft tasks. 
MLSH \citep{frans2017mlsh} pretrains a hierarchy with separate low-level policies on easier tasks, alternating update phases between the two levels. 
HeLMS \citep{rao2021helms} learns reusable robot manipulation skills from a given diverse dataset. 
MODAC \citep{veeriah2021modac} uses meta gradients to learn low-level policies that are helpful for solving tasks. 
However, all these approaches require manually specifying a diverse distribution of training tasks, and it is unclear whether generalization beyond the training tasks will be possible. 
Instead of relying on task rewards for learning skills, Director learns the worker as a goal-conditioned policy with a dense similarity function in feature space.

\paragraph{Mutual information}

Mutual information approaches allow automatic discovery of skills that lead to future states. 
VIC \citep{gregor2016vic} introduced a scalable recipe for discovering skills by rewarding the worker policy for reaching states from which the latent skill can be accurately predicted, effectively clustering the trajectory space. 
Several variations of this approach have been developed with further improvements in stability and diversity of skills, including SSN4HRL \citep{florensa2017snn}, DIAYN \citep{eysenbach2018diayn} and VALOR \citep{achiam2018valor}. 
DISCERN \citep{warde2018discern} and CIC \citep{laskin2022cic} learn a more flexible similarity function between skills and states through contrastive learning. 
DADS \citep{sharma2019dads} estimates the mutual information in state-space through a contrastive objective with a learned model. 
DISDAIN overcomes a collapse problem by incentivizing exploration through ensemble disagreement \citep{strouse2021disdain}. 
LSP \citep{xie2020lsp} learns a world model to discover skills that have a high influence on future representations rather than inputs. 
While these approaches are promising, open challenges include learning more diverse skills without dropping modes and learning skills that are precise enough for solving tasks.

\paragraph{Goal reaching}

Learning the low-level controller as a goal-conditioned policy offers a stable and general learning signal. 
HER \citep{andrychowicz2017her} popularized learning goal-conditioned policies by combining a sparse reward signal with hindsight, which HAC \citep{levy2017hac} applied to learn an interpretable hierarchy with three levels by manually designing task-relevant goal spaces for each task. 
Instead of relying on hindsight with a sparse reward, HIRO \citep{nachum2018hiro} employs a dense reward in observation space based on the L2 norm and solves Ant Maze tasks given privileged information. 
NORL \citep{nachum2018norl} introduces a representation learning objective to learn a goal space from a low-resolution top-down image instead but still uses a dense ground-truth reward for learning the manager. 
HSD-3 \citep{gehring2021hsd3} also learns a hierarchy with three levels and uses the robot joint space combined with a mask to specify partial goals. 
While not learning a high-level policy, RIG \citep{nair2018rig} computes goal rewards in the latent space of an autoencoder, allowing them to reach simple visual goals. 
FuN \citep{vezhnevets2017fun} provides task and goal rewards to its low-level policy and is trained from pixels but provides limited benefits over a flat policy trained on task reward. 
SecTAR \citep{co2018sectar} learns a sequence autoencoder that serves to propose goals and compute distances for low-dimensional environments and shows fast learning in low-dimensional environments with sparse rewards by high-level planning. 
DDL \citep{hartikainen2019ddl} learns to reach goals from pixels by learning a temporal distance function. 
LEXA \citep{mendonca2021lexa} learns a goal conditioned policy inside of a world model that achieves complex multi-object goal images, but assumes goals to be specified by the user. 
By comparison, Director solves difficult sparse reward tasks end-to-end from pixels without requiring domain-specific knowledge by learning a world model.

\clearpage
\section{Full Visual Control Suite}
\label{sec:dmc}
\begin{figure}[h!]
\centering
\includegraphics[width=0.98\textwidth]{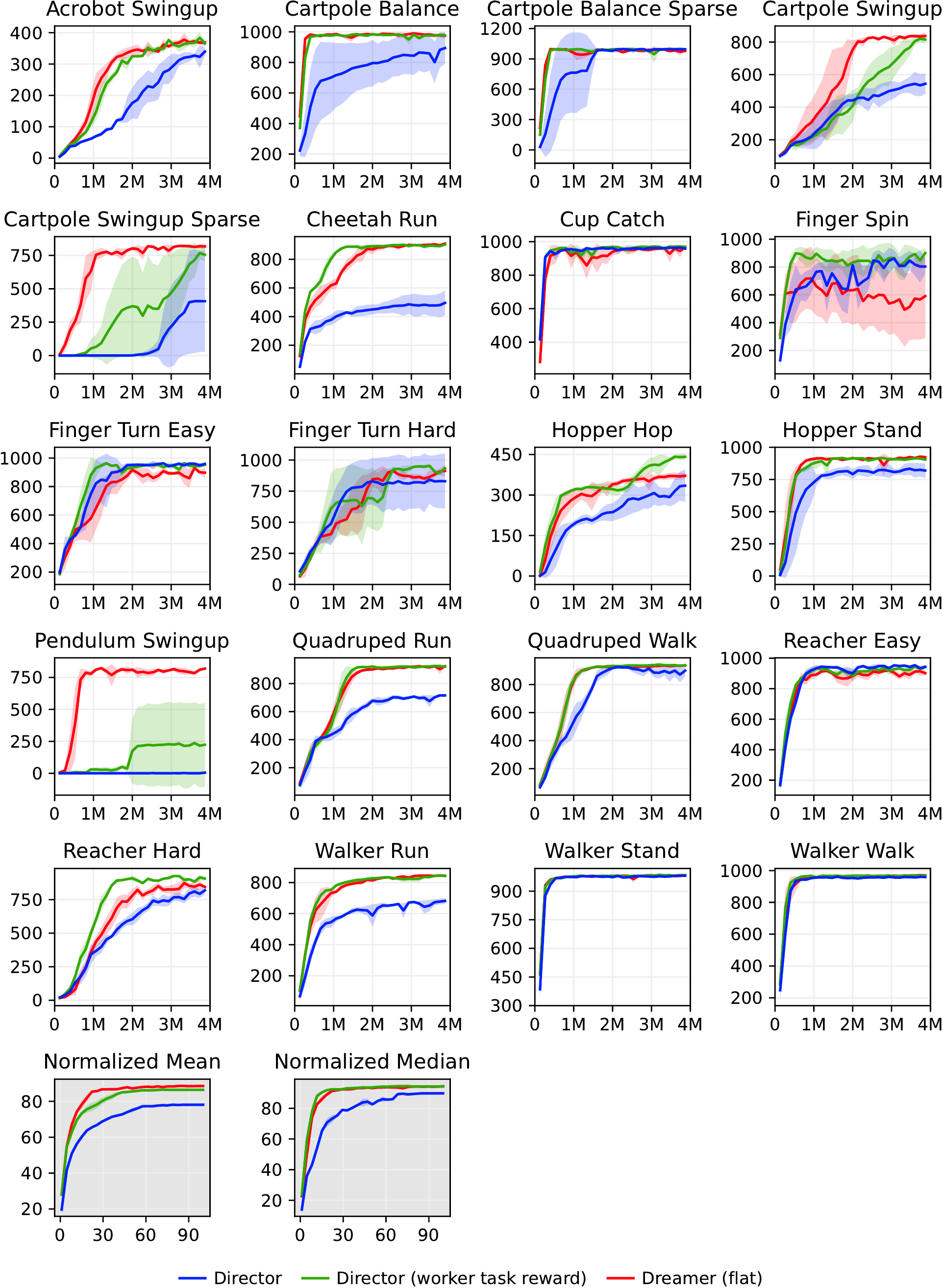}
\caption{
To test the generality of Director, we evaluate it on a diverse set of 20 visual control tasks \citep{tassa2018dmcontrol} without action repeat.
We find that Director solves a wide range of tasks despite giving no task reward to the worker, for the first time in the literature demonstrating a hierarchy with task-agnostic worker performing reliably across many tasks. 
When additionally providing task reward to the worker, performance reaches the state-of-the-art of Dreamer and even exceeds it on some tasks.
These experiments use no action repeat and train every 16 environment steps, resulting in faster wall clock time and lower sample efficiency than the results reported by \citet{hafner2019dreamer}.
}
\label{fig:dmc}
\end{figure}

\clearpage
\section{Full Atari Suite}
\label{sec:atari}
\begin{figure}[h!]
\centering
\includegraphics[width=0.97\textwidth]{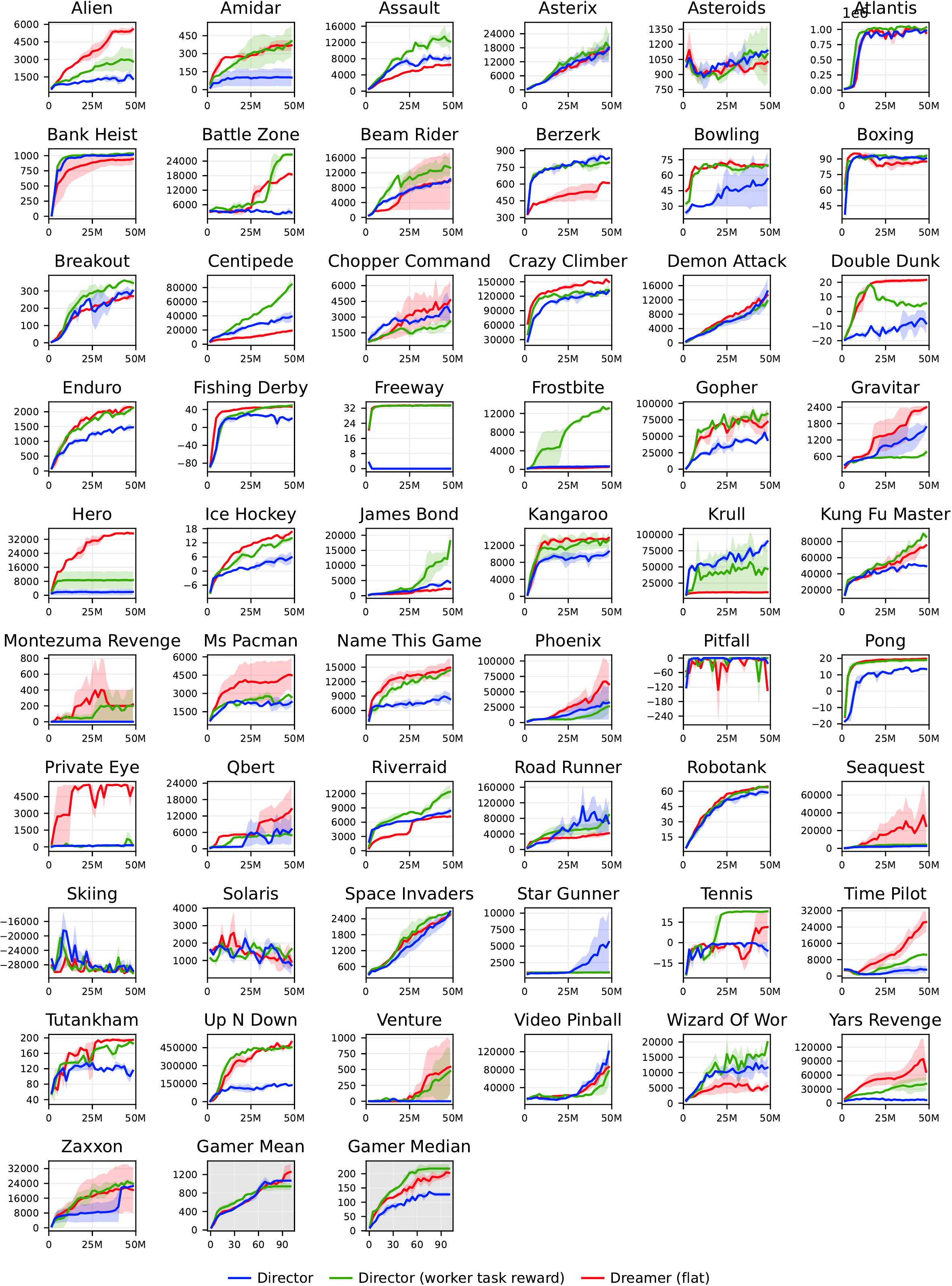}
\caption{
To test the generality of Director, we evaluate it on 55 Atari games \citep{bellemare2013ale}. 
We find that Director solves a wide range of tasks despite giving no task reward to the worker.
Director is the first approach in the literature that demonstrates a hierarchy with task-agnostic worker performing reliably across many tasks. 
When additionally providing task reward to the worker, performance reaches that of Dreamer and even exceeds its human-normalized median score.
These experiments use no action repeat and train every 16 environment steps, resulting in faster wall clock time and lower sample efficiency than the results reported by \citet{hafner2019dreamer}.
}
\label{fig:atari}
\end{figure}

\clearpage
\section{Additional Goal Visualizations}
\label{sec:moregoals}
\begin{figure}[h!]
\centering

\begin{subfigure}{\textwidth}
\includegraphics[width=\textwidth]{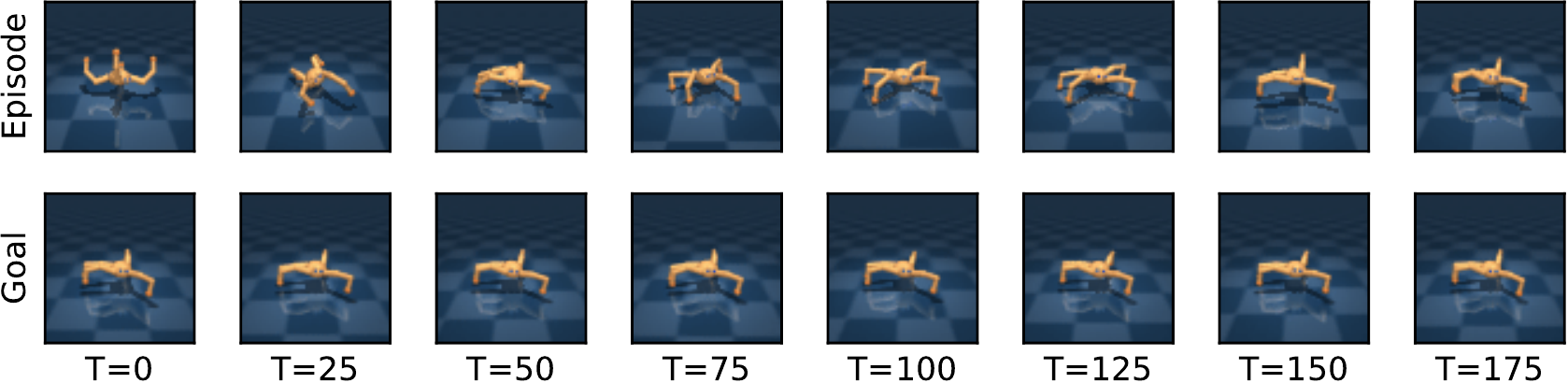}
\caption{
Quadruped Walk. 
The manager learns to abstract away leg movements by requesting a forward leaning pose with shifting floor pattern, and the worker fills in the leg movements. 
The underlying latent goals are Markov states and thus likely contain a description of high forward velocity.
}
\end{subfigure} \\[2ex]

\begin{subfigure}{\textwidth}
\includegraphics[width=\textwidth]{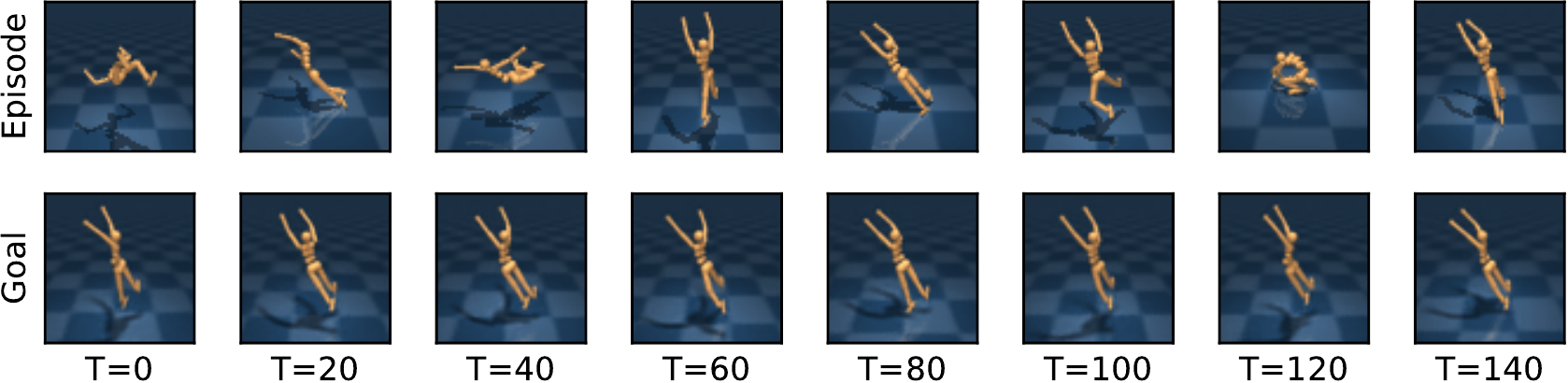}
\caption{
Humanoid Walk. 
The manager learns to direct the worker via an extended pose with open arms, which causes the worker to perform energetic forward jumps that are effective at moving the robot forward. 
While the visualizations cannot show this, the underlying latent goals are Markovian and can contain velocity information.
}
\end{subfigure} \\[2ex]

\begin{subfigure}{\textwidth}
\includegraphics[width=\textwidth]{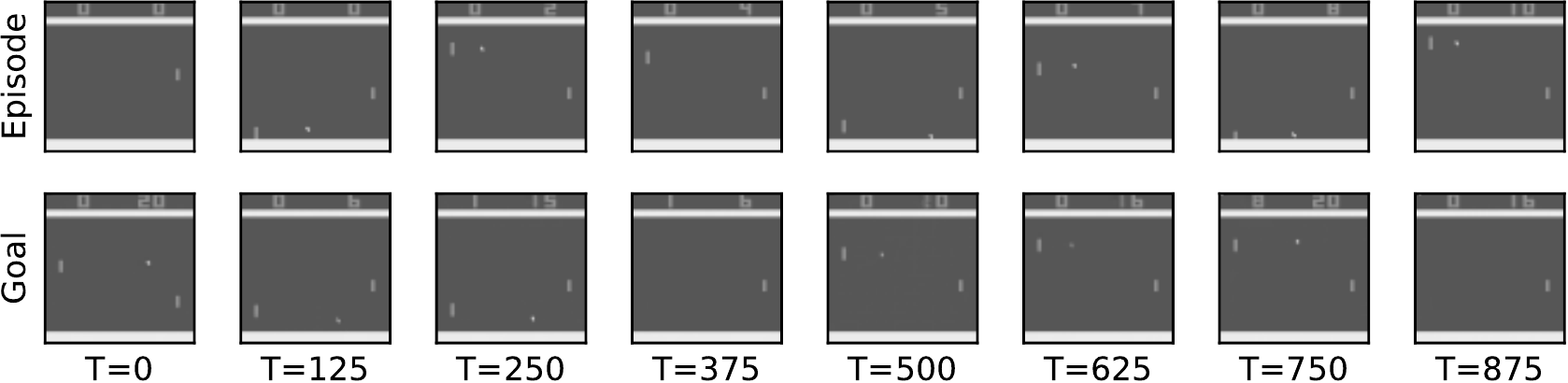}
\caption{
Pong. 
The manager directs the worker simply by requesting a higher score via the score display at the top of the screen. 
The worker follows the command by outplaying the opponent. 
We find that for tasks that are easy to learn for the worker, the manager frequently chooses this hands-off approach to managing.
}
\end{subfigure} \\[2ex]

\begin{subfigure}{\textwidth}
\includegraphics[width=\textwidth]{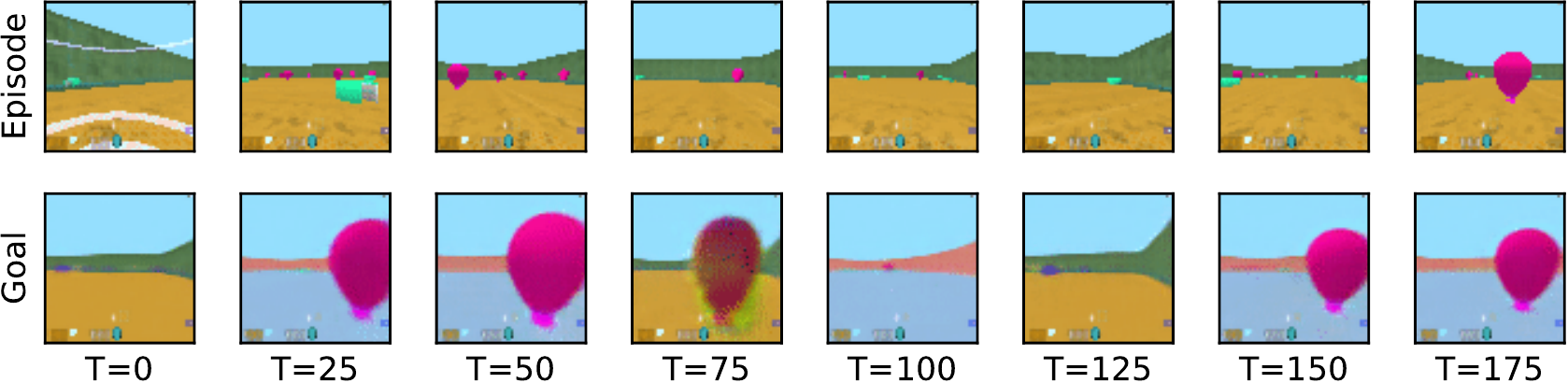}
\caption{
DMLab Collect Good. 
The manager steers through the arena by targeting the good objects and avoiding the bad ones. 
The worker succeeds at reaching the balloons regardless of the wall textures, showing that it learns to focus on achievable aspects of the goals without being distracted by non-achievable aspects.
}
\end{subfigure}

\caption{Additional goal visualizations.}
\label{fig:more}
\end{figure}

\end{document}